\documentclass[11pt]{article}

\usepackage[final]{acl}

\usepackage{times}
\usepackage{latexsym}

\usepackage[T1]{fontenc}

\usepackage[utf8]{inputenc}

\usepackage{microtype}

\usepackage{inconsolata}

\usepackage{graphicx}
\usepackage{amsmath}
\usepackage{amssymb}
\usepackage{booktabs}
\usepackage{multirow}
\usepackage[table]{xcolor}
\definecolor{lightgray}{gray}{0.9}

\usepackage{subcaption}
\usepackage{graphicx}
\usepackage[most]{tcolorbox}
\tcbuselibrary{breakable}
\usepackage{kotex}

\newcommand{\bd}{\textperiodcentered}   
\newcommand{\tk}[1]{\texttt{#1}}         

\title{Steering Language Models Before They Speak: Logit-Level Interventions}

\newtcolorbox{promptbox}[1][]{
  colback=gray!5,
  colframe=gray!80!black,
  coltitle=gray!30!black,
  colbacktitle=gray!30,
  boxrule=0.8pt,
  arc=2pt,
  left=6pt,
  right=6pt,
  top=4pt,
  bottom=4pt,
  fonttitle=\bfseries,
  title=Prompt,
  breakable = true,
  #1
}

\newtcolorbox{rqbox}{
  enhanced,
  colback=brown!6,            
  colframe=brown!65!black,
  boxrule=0.6pt,
  arc=2pt,
  left=6pt, right=6pt, top=5pt, bottom=5pt,
}

\author{
  Hyeseon An$^1$,
  Shinwoo Park$^{2}$\thanks{This work was carried out while the author was at Yonsei University.},
  Hyundong Jin$^1$,
  \and
  Yo-Sub Han$^{1}$\thanks{Corresponding Author.} \\
  $^1$Yonsei University \\
  $^2$University of Luxembourg \\
  \texttt{\{\href{mailto:hsan@yonsei.ac.kr}{hsan},
  \href{mailto:tuzi04@yonsei.ac.kr}{tuzi04},
  \href{mailto:emmous@yonsei.ac.kr}{emmous}\}@yonsei.ac.kr}, \\
  \href{mailto:shinwoo.park@uni.lu}{\texttt{shinwoo.park@uni.lu}}
}

\begin{document}
\maketitle
\begin{abstract}
Controllable generation requires language models to realize output characteristics such as reading level, politeness, and toxicity. Existing steering methods are often indirect, require access to internal activations, or depend on auxiliary trained models. We propose \texttt{SWAI}, a training-free inference-time method that addresses these limitations by steering directly in logit space using corpus-derived token statistics. \texttt{SWAI} computes z-normalized one-vs-rest log-odds scores from labeled corpora and biases high-scoring tokens only within the model's top-$K$ candidate set, allowing control to favor target-characteristic tokens while preserving contextually plausible choices. Across readability, politeness, and toxicity control, \texttt{SWAI} consistently improves over prompt-based and prior logit-level baselines without modifying model parameters, accessing internal layers, or training an auxiliary model. Selectivity and lookup-table ablations show that the gains come from target-specific statistical scores rather than generic logit perturbation. These results indicate that effective steering does not require learned controllers when the logit intervention is guided by target-specific statistics under high-probability candidates. Code is available at \url{https://github.com/hsannn/swai}.

\end{abstract}

{\noindent\textcolor{red}{\textbf{Warning}: this paper contains content that may be offensive and upsetting.}}

\section{Introduction}
\label{sec:intro} 

Real-world uses of large language models (LLMs) rarely require text for its own sake;
they require text that satisfies particular output requirements \citep{zhang2023survey, zhou2023ctg}.
Such text may need to match a target reading level for readers with different
literacy levels~\citep{vajjala2018onestopenglish}, express appropriate politeness in user-facing
interactions~\citep{danescu2013computational}, or avoid toxic language in safety-sensitive use~\citep{gehman2020realtoxicityprompts, schick2021self}.
We refer to these requirements as \emph{output characteristics}: properties that concern not what a model writes about, but how it writes.
However, these output characteristics are not reliably realized by simply specifying them in a prompt \citep{jie-etal-2024-prompt, zhou2023ctg}, since prompts adjust only the model's input.
This limitation has motivated extensive work on controllable text generation (CTG),
which seeks to align model outputs with target output characteristics without sacrificing fluency.
Within this space, a broad family of \emph{steering} methods seeks to control a pretrained model toward
target output characteristics, and they differ in how and where they intervene.

Prompt-based steering~\citep{jie-etal-2024-prompt} is simple and
architecture-agnostic, but the control signal is indirect. The model may appear to follow
the instruction while failing to satisfy the requested property in the generated
text \citep{jie-etal-2024-prompt}.
Training-based steering instead learns control behavior through model parameters, using methods
such as control codes~\citep{keskar2019ctrl}, prefix tuning~\citep{li-liang-2021-prefix}, or instruction tuning~
\citep{zhou2023ctg}. This yields strong and
stable control, but at the cost of labeled
data and computation, and the learned behavior is bound to one trained model.
Activation-based steering~\citep{Dathathri2020Plug, turner2023steering, rimsky2024steering} edits internal
hidden states at inference time, adding a steering direction to the residual stream.
It avoids retraining, but requires architecture-specific layer selection and may
degrade fluency or reasoning by perturbing internal representations.

Notably, most existing steering methods do not directly intervene in the logit space
where next-token probabilities are finalized. 
This space is nevertheless a natural target for controllability, since output characteristics are
ultimately realized through the token choices sampled from the final decoding
distribution \citep{li2023contrastive}. 
A small number of methods operate at the logit level, demonstrating that the decoding distribution is an effective site of control.
\citet{krause-etal-2021-gedi-generative} reweights candidate generations using a discriminator,
and \citet{liu-etal-2021-dexperts} contrasts a trained auxiliary model with an untrained model,
with both methods steering generation by reshaping the base model's output logits.
However, these methods typically
derive their control signals from auxiliary trained models. As a result, their
effectiveness is entangled with the cost and model dependence introduced by
auxiliary training.
This raises the following simpler yet crucial research question.

\begin{rqbox}
    \textbf{RQ.} Can a lightweight control signal from corpus statistics steer a model effectively while preserving generation quality,
    without any additional training?
\end{rqbox}

Building on this view, we propose \textbf{\texttt{SWAI}}
(\textbf{S}tatistical \textbf{W}riting style \textbf{A}ligned \textbf{I}nference), a
training-free, inference-time logit intervention. \texttt{SWAI} precomputes a token-level score
table from labeled corpora to quantify how strongly each token characterizes a target output characteristic.
At each decoding step, \texttt{SWAI} uses these scores to identify favored tokens and adds a constant bias to their logits, shifting the output toward the target characteristic without modifying the model's parameters or internal representations.

Experiments confirm that
\texttt{SWAI} consistently outperforms prompt-based and prior logit-level baselines, without any training.
We also show effectiveness on safety-critical toxicity control,
demonstrating that a single mechanism transfers across qualitatively different objectives.
Taken together, these results position logit-level steering from corpus statistics as
a practical and broadly applicable approach to controllable generation.

Our contributions are summarized as follows:
\begin{itemize}
\item We construct a token-level score table that quantifies output characteristics,
precomputed offline for constant-time lookup at decoding.
\item We introduce a training-free steering method that biases the base model's
output logits with these scores at inference time, without any auxiliary model or
layer-specific tuning.
\item We evaluate \texttt{SWAI} on reading level, politeness, and toxicity, and show
through analysis that the control reflects genuine steering rather than source-label
copying.
\end{itemize}

\section{Related Work}
\label{sec:related_work}
Steering methods for LLMs can be distinguished by where they intervene
during generation.

\subsection{Training-based Steering}
Training-based steering incorporates controllability into model parameters through additional training.
\citet{keskar2019ctrl} trained a model from scratch on text prefixed with tokens that specify
a desired domain or style. Rather than retraining the entire model, \citet{li-liang-2021-prefix}
froze the backbone and learned only a small continuous vector prepended to the input.
Others left the architecture intact and shaped behavior through targeted training signals.
\citet{zhou2023ctg} instruction-tuned on weakly supervised constraint-instruction pairs,
\citet{de-langis-etal-2024-dynamic} jointly optimized multiple style rewards while balancing their weights,
and \citet{wang-etal-2025-verifiable} trained on programmatically checkable format data that the model can verify.
Although effective, these methods require labeled data and additional computation,
and the learned control remains tied to the trained model.

\subsection{Activation-based Steering}

Activation-based steering guides a frozen model by intervening on its internal
activations at inference time, without updating model weights. Early work
typically constructed a steering direction from contrastive activation patterns
and injects it during generation. \citet{turner2023steering} used activation
differences between opposing prompts, while \citet{rimsky2024steering} averaged
such differences across many examples to obtain a more stable direction. Other
methods derived steering directions from concepts or tasks rather than prompt
pairs, either by extracting leading principal components of activations under
contrastive stimuli as interpretable concept directions
\citep{zou2023representation} or by distilling task signals from in-context
examples into vectors that elicit the corresponding task when injected
\citep{todd2024function,liu2024context}. Later work refined these interventions.
Some methods moved beyond direct vector addition by transforming activations
while preserving their magnitude \citep{pham2024householder} or by projecting
out directions associated with undesired behavior \citep{qiu2024spectral}.
Others specialized activation interventions for alignment-oriented goals,
including instruction adherence \citep{stolfoimproving}, robust knowledge
editing \citep{scialanga-etal-2025-sake}, hallucination suppression
\citep{su2025activation}, and safe responses without blanket refusal
\citep{ghosh2025simple}. These methods avoided retraining, but they depend on
architecture-specific choices of layer and direction, and perturbing internal
representations can degrade fluency or reasoning.

\begin{figure*}[t!]
    \centering
    \includegraphics[width=\textwidth]{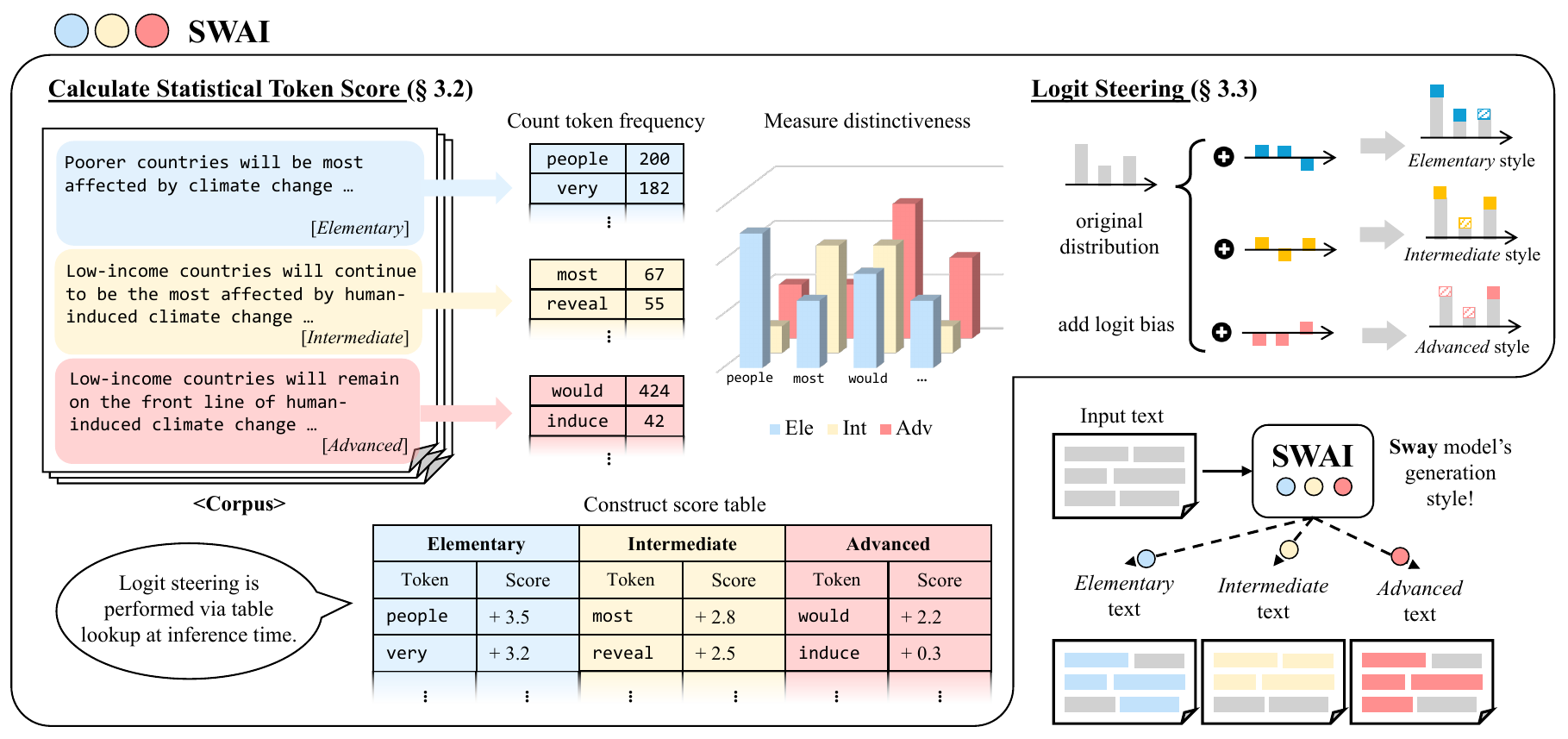}
    \caption{Overview of \texttt{SWAI}. It first derives token-level statistical scores from labeled corpora
via smoothed one-vs-rest log-odds with variance normalization.
During decoding, intervention is confined to a top-$K$ candidate set,
where a fixed logit offset is applied to the highest-scoring tokens
before sampling from the modified distribution.}
    \label{fig:overview}
\end{figure*}

\subsection{Logit-level Steering}
Logit-level steering intervenes directly in logit space by reshaping the next-token distribution
\citep{kumar2021controlled, kumar2022gradient}.
\citet{krause-etal-2021-gedi-generative} reweighted the base distribution with a
class-conditional language model used as a discriminator, and
\citet{liu-etal-2021-dexperts} added the logit difference between an expert and
an anti-expert model. \citet{yang-klein-2021-fudge} reweighted outputs using a
discriminator that predicts whether a partial sequence will satisfy the target
attribute, while \citet{meng2022nado} decomposed sequence-level oracle feedback
into token-level guidance~\citep{gu2022improving}. Although effective, these methods rely on 
auxiliary models for their control signals \citep{deng2023reward},
introducing an additional training requirement.

\section{Methodology}
\label{sec:method}
We propose \textbf{\texttt{SWAI}},
a lightweight and training-free logit steering framework
designed to guide the output characteristics of LLMs through inference-time intervention. \texttt{SWAI}
recalibrates the model's
output distribution by utilizing token-level statistics derived from labeled corpora.
The framework consists of two primary phases:
(1)~the offline construction of a statistical token score table and
(2)~the execution of logit steering during the autoregressive decoding process.

\subsection{Preliminaries}
\label{subsec:prelim}
A standard autoregressive language model $f_{\theta}$ generates a logit vector $z_t \in \mathbb{R}^{|V|}$ for the
current time step $t$, given the previously generated sequence $x_{<t}$, where $|V|$ denotes the size of the vocabulary:
\begin{equation*}
    z_t = f_{\theta}(x_{<t}).
\end{equation*}

The logit vector $z_t$ is transformed into a probability distribution $P_t$ over each token $v \in V$ using the
softmax function:
\begin{equation*}
    P_t(v) = \frac{\exp(z_t(v))}{\sum_{u \in V} \exp(z_t(u))}.
\end{equation*}

In standard decoding, the model samples the next token $x_t$ from this distribution $P_t$.
\texttt{SWAI} intervenes at the logit level, specifically after the generation of $z_t$ but prior to the softmax transformation,
to perform logit steering toward a target output characteristics.

\subsection{Statistical Token Score}
\label{subsec:scoring}
We formalize the relationship between tokens and specific output characteristics by constructing a statistical token score table.
This scoring mechanism quantifies the statistical association between each token $v$ and a target output characteristic
$r$ through a comparative analysis of labeled corpora.

\paragraph{Log-odds Ratio Calculation.}
We calculate the log-odds ratio between a corpus possessing output characteristic $r$ and a contrastive corpus
$\neg r$ to measure the distinctiveness of token $v$ for output characteristic $r$. We define the statistics
for the contrastive corpus as follows:
\begin{itemize}
    \item $c_{\neg r}(v) = \sum_{r' \neq r} c_{r'}(v)$: The total frequency of token $v$ across all classes other than $r$.
    \item $N_{\neg r} = \sum_{u \in V} c_{\neg r}(u)$: The total number of tokens in the contrastive corpus.
\end{itemize}

Data sparsity and numerical instability are mitigated by incorporating a Dirichlet prior $\alpha_v$ for smoothing.
In our implementation, we adopt a pooled Dirichlet prior by defining aggregated token counts
$c_{\text{pooled}}(v)=c_r(v)+c_{\neg r}(v)$ and $\pi_v=\max\!\left(1, c_{\text{pooled}}(v)\right)$.
We then set the token-specific prior mass as $\alpha_v=\alpha\cdot\pi_v$ with $\alpha=0.01$, and define
$\alpha_0=\sum_{u\in V}\alpha_u$ accordingly.
We formulate the token score $s_r(v)$ for each token $v$ as:
\begin{equation}
\label{eq:score}
\begin{aligned}
s_r(v)
&= \log \frac{c_r(v)+\alpha_v}{(N_r+\alpha_0)-\left(c_r(v)+\alpha_v\right)} \\
&\quad - \log \frac{c_{\neg r}(v)+\alpha_v}
{(N_{\neg r}+\alpha_0)-\left(c_{\neg r}(v)+\alpha_v\right)},
\end{aligned}
\end{equation}

where $c_r(v)$ represents the frequency of token $v$ in the output characteristic $r$ corpus, $N_r$ is the total token count of
the corpus, and $\alpha_0 = \sum_{u \in V} \alpha_u$ denotes the sum of the prior mass.

\paragraph{Intuition.}
The first log term in Eq.~(\ref{eq:score}) measures the smoothed odds of token $v$ in the target class $r$, whereas the second measures its corresponding odds in the contrastive corpus $\neg r$. Their difference is therefore positive when $v$ is disproportionately associated with the target class, close to zero when it occurs similarly across the target and contrastive classes, and negative when it is more characteristic of the contrastive classes. The Dirichlet prior stabilizes these odds for sparse token counts, preventing rare observations from producing excessively large scores.

\paragraph{Z-score Normalization.}
Raw log-odds ratios can be susceptible to noise from low-frequency tokens.
We address this by applying $z$-score normalization based on the estimated variance of the log-odds ratio:
\begin{equation}
\begin{aligned}
z_r(v)
&= \frac{s_r(v)}{\sqrt{\sigma^2\!\left(s_r(v)\right)}}, \\
\sigma^2\!\left(s_r(v)\right)
&\approx \frac{1}{c_r(v)+\alpha_v}
      + \frac{1}{c_{\neg r}(v)+\alpha_v}.
\end{aligned}
\end{equation}

The denominator functions as the standard error, which suppresses the scores of low-frequency tokens.
The resulting $z_r(v)$ provides a standardized metric of statistical confidence, indicating the
degree to which a token characterizes the target output characteristic $r$.
Unlike TF-IDF, which measures token importance relative to documents, our scoring function explicitly
captures class-conditional asymmetry via a one-vs-rest log-odds formulation, yielding a directional signal
suitable for decoding-time output characteristic control.

\paragraph{Construction of the Statistical Score Table.}
The final output of this scoring phase is the statistical score table, a precomputed mapping that associates
each token $v$ with its corresponding $z$-score $z_r(v)$. 
This static table enables efficient, constant-time lookup during steering, 
allowing \texttt{SWAI} to retrieve the statistical significance of any candidate token without 
additional online computation.

\subsection{Logit Steering}
\label{subsec:steering}
During the decoding phase, \texttt{SWAI} performs logit steering by recalibrating the output distribution
at each time step $t$. 
This intervention reinforces the target output characteristic while preserving semantic integrity of the base model.

\paragraph{Candidate Filtering.}
At each step $t$, we construct a candidate set $C_t$ from the original logits $z_t$ using filtering methods
top-$k$ sampling. This constraint ensures that the logit steering is restricted to tokens
that the model considers contextually plausible, thereby preserving linguistic fluency.

\paragraph{Targeted Logit Bias.}
From the candidate set $C_t$, we identify a favored set $F_t$ containing the top $m$ tokens with the highest
$z_r(v)$ scores:
\begin{equation}
    F_t = \{ v \mid v \in C_t \text{ and } \text{rank}(z_r(v)) \leq m \}.
\end{equation}

A constant bias $\delta$ is applied to the original logits of the tokens in $F_t$, steering the
model toward the desired output characteristic and producing the adjusted logits $z'_t$.
\begin{equation}
    z'_t(v)=\begin{cases}z_t(v)+\delta, & \text{if } v\in F_t \\z_t(v), & \text{otherwise}\end{cases}
\end{equation}

\begin{table*}[h!]
\centering
\begin{tabular}{l|l|ccccccc}
\toprule
Model & Method & Acc & $F_1$ & Precision & Recall & Perplexity & $\kappa$ & MCC \\
\midrule
\rowcolor{lightgray} \multicolumn{9}{c}{\textsc{OSE} (Rewriting)} \\

\multirow{4}{*}{Llama3.1 8B}
 & Prompting & 37.11\% & 0.363 & 36.20\% & 36.48\% & 11.99 & 0.052  & 0.052 \\
 & GeDi & 38.96\% & 0.392 & 41.14\% & 39.16\% & 11.88 & 0.091  & 0.093 \\
 & DExperts & 35.06\% & 0.173 & 33.33\% & 11.69\% & 13.12 & 0.000 & 0.000 \\
 & \texttt{SWAI} (ours) & \textbf{84.53\%} & \textbf{0.845} & \textbf{84.53\%} & \textbf{84.53\%} & \textbf{8.88} & \textbf{0.768} & \textbf{0.768} \\
\midrule
\multirow{4}{*}{Llama3.2 1B}
 & Prompting & 28.87\% & 0.289 & 29.36\% & 28.99\% & 13.49 & -0.063 & -0.064 \\
 & GeDi & 37.66\% & 0.369 & 40.69\% & 39.24\% & 13.46 & 0.076 & 0.082 \\
 & DExperts & 15.58\% & 0.090 & 33.33\% & 5.19\% & 14.44 & 0.000 & 0.000 \\
 & \texttt{SWAI} (ours) & \textbf{71.11\%} & \textbf{0.712} & \textbf{71.27\%} & \textbf{71.11\%} & \textbf{12.48} & \textbf{0.567} & \textbf{0.567} \\
\midrule
\rowcolor{lightgray} \multicolumn{9}{c}{\textsc{WikiPol}} \\

\multirow{4}{*}{Llama3.1 8B}
 & Prompting & 56.70\% & 0.569 & 56.79\% & 57.10\% & 18.58 & 0.351 & 0.351 \\
 & GeDi & 22.08\% & 0.217 & 46.13\% & 14.62\% & 51.97 & 0.083 & 0.148 \\
 & DExperts & 68.83\% & 0.628 & 62.49\% & 63.31\% & \textbf{15.62} & 0.412 & 0.414 \\
 & \texttt{SWAI} (ours) & \textbf{77.20\%} & \textbf{0.730} & \textbf{82.89\%} & \textbf{69.28\%} & 25.92 & \textbf{0.594} & \textbf{0.626} \\
\midrule
\multirow{4}{*}{Llama3.2 1B}
 & Prompting & 49.48\% & 0.494 & 49.34\% & 49.59\% & 38.81 & 0.241 & 0.241 \\
 & GeDi & 15.58\% & 0.150 & 28.76\% & 10.32\% & 64.73 & -0.017 & -0.027 \\
 & DExperts & 71.43\% & 0.582 & 64.66\% & 56.67\% & \textbf{21.70} & 0.474 & 0.480 \\
 & \texttt{SWAI} (ours) & \textbf{79.20\%} & \textbf{0.709} & \textbf{84.96\%} & \textbf{66.37\%} & 31.85 & \textbf{0.582} & \textbf{0.616} \\
\bottomrule
\end{tabular}
\caption{Main results on \textsc{OSE} and \textsc{WikiPol} datasets. Across both datasets, \texttt{SWAI} consistently outperforms prompt-based baselines.}
\label{tab:main}
\end{table*}

\paragraph{Characteristic-guided Sampling.}
We transform the steered logit vector $z'_t$ into a modified probability distribution $P'_t$ via the softmax function:
\begin{equation}
    P_t'(v) = \frac{\exp(z_t'(v))}{\sum_{u \in V} \exp(z_t'(u))}.
\end{equation}

By sampling the next token $x_t$ from $P'_t$, the model becomes statistically more likely to select tokens aligned to the desired characteristic while adhering to the grammatical constraints dictated by the original distribution.

\section{Experimental Results}
\label{sec:experiment}
\subsection{Setup}
We evaluate SWAI on three datasets, each targeting a distinct output characteristic: \textsc{OSE} \citep{vajjala2018onestopenglish} for reading level, \textsc{WikiPol} \citep{danescu2013computational} for politeness, and \textsc{RealTox} \citep{gehman2020realtoxicityprompts} for toxicity.
For logit steering, we use the base (non-instruction-tuned) variants of Meta's
Llama models, specifically Llama3.1 8B and Llama3.2 1B. Both models are used
as pretrained base language models without instruction tuning. Text generation
is performed with identical decoding settings for both models, using a
temperature of 0.8, top-p of 0.95, and an n-gram repetition penalty of 3. For
evaluation, we employ GPT-4o from OpenAI as a judge LLM, with the temperature
fixed to 0.0 to ensure deterministic and consistent judgments. This
configuration allows for controlled diversity during generation while
maintaining stable and reliable evaluation.
Further details on the datasets, hyperparameters, and evaluation metrics are
provided in Appendix~\ref{app:details}, and the full system prompts given to
the judge LLM are listed in Appendix~\ref{app:prompt}.

\subsection{Readability and Politeness Control}
\label{subsec:results}
The results in Table~\ref{tab:main} clearly highlight the
fundamental differences between prompt-based control and the proposed
statistics-driven logit steering approach, \texttt{SWAI}. The prompt-only
baseline exhibits low accuracy and $F_1$ scores across both datasets, and
performs close to random on \textsc{OSE} in particular. Despite this, the
baseline achieves relatively high confidence scores, suggesting that
prompt-based steering can induce `confidently wrong' generations, where the
model appears strongly aligned with the prompt while failing to realize the
target output characteristic. In contrast, \texttt{SWAI} achieves substantial improvements
across Accuracy, $F_1$, $\kappa$, and MCC without any additional training or
fine-tuning, providing strong quantitative evidence that the generated text is
consistently shifted toward the desired output characteristic.

Interestingly, while \texttt{SWAI} substantially improves performance on
\textsc{OSE} without materially changing confidence scores, it leads to lower
confidence on \textsc{WikiPol} despite significant gains in accuracy and
reliability metrics. This pattern suggests that \texttt{SWAI} may avoid
over-reliance on rigid or exaggerated surface cues, instead expressing output characteristics
in a more natural and diverse manner that reduces the judge model's certainty.
Furthermore, the comparison across model sizes indicates that sufficient model
capacity is important for maintaining stable steering effects in output characteristics
requiring long-range consistency, such as reading difficulty, whereas output characteristics
like politeness can be effectively controlled even with smaller models.

\begin{figure}[t]
    \centering
    \includegraphics[width=\linewidth]{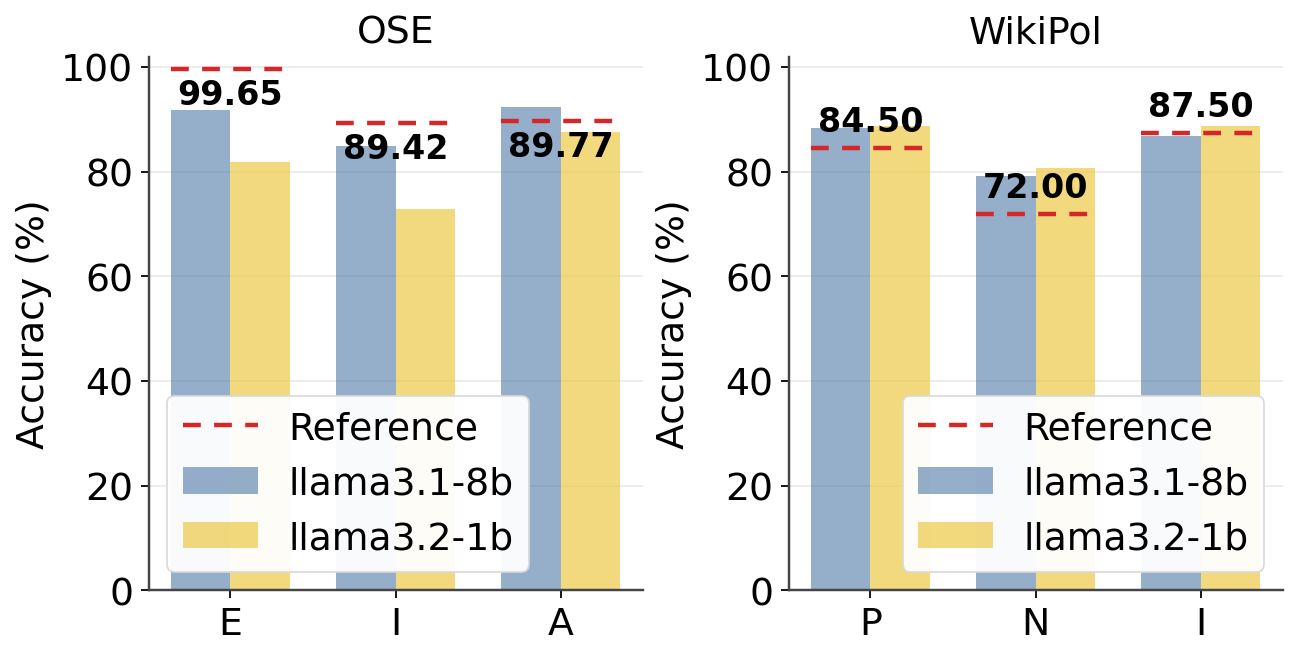}
    \caption{Class-wise Accuracy on \textsc{OSE} and \textsc{WikiPol} datasets.
    Reference denotes the Judge LLM performance on the original data before logit steering.}
    \label{fig:class}
\end{figure}

\paragraph{Class-wise Results.}
Figure~\ref{fig:class} reports class-wise accuracy on \textsc{OSE} and \textsc{WikiPol}, comparing
logit-steered generations against a reference score obtained by applying the LLM
judge directly to the original dataset without steering (red dashed lines). This
reference serves as an approximate upper bound, reflecting the judge model's
ability to separate classes on naturally occurring, human-written data. Here,
E/I/A in \textsc{OSE} denote Elementary, Intermediate, and Advanced, while P/N/I in
\textsc{WikiPol} correspond to Polite, Neutral, and Impolite, respectively. Across both
datasets, logit steering achieves accuracy comparable to the
reference across all classes, and in several cases matches or exceeds the
reference performance. On \textsc{OSE}, strong separability is maintained across all
difficulty levels, while on \textsc{WikiPol}, steering notably improves performance on
the Neutral class, which is typically more ambiguous. These results demonstrate
that decoding-time logit steering preserves class-level separability, and can
even yield cleaner output characteristic realization than the original data, despite
requiring no additional training.

\subsection{Toxicity Control}
\label{subsec:toxicity}
\textsc{RealTox} provides a safety-critical test case in which control is complicated
by sparsity, class imbalance, and contextual toxicity cues. As shown in
Table~\ref{tab:realtox}, the prompt-only baseline achieves 100\% precision but
only 0.50\% recall, yielding an $F_1$ score near zero. This pattern indicates an
avoidance-style collapse: the model rarely produces outputs judged as toxic, but
therefore fails to realize controlled toxic continuations when required. In this
setting, accuracy alone is misleading, making recall and $F_1$ essential for
evaluating toxicity control.

By contrast, \texttt{SWAI} improves recall while retaining high precision, leading
to the best accuracy and a substantially higher $F_1$ score than prompting. These
results show that the same statistics-driven logit intervention extends beyond
readability and politeness to a sparse safety-related attribute, without
additional training or fine-tuning.

\begin{table}[t]
\centering
\resizebox{\linewidth}{!}{
\begin{tabular}{lcccc}
\toprule
Method & Acc. & $F_1$ & Prec. & Recall \\
\midrule
Prompting & 50.25\% & 0.010 & \textbf{100.00\%} & 0.50\% \\
GeDi & 52.56\% & 0.461 & 54.92\% & 52.56\%   \\
DExperts & 76.92\% & \textbf{0.585} & 72.22\% & \textbf{58.28\%} \\
\texttt{SWAI} \textit{(ours)} & \textbf{78.33\%} & 0.552 & 97.56\%  & 38.46\% \\
\bottomrule
\end{tabular}}
\caption{Results on the \textsc{RealTox} dataset.
Performance is reported for prompt-based toxicity control and \texttt{SWAI}.}
\label{tab:realtox}
\end{table}

\begin{figure}[t]
    \centering
    \includegraphics[width=\linewidth]{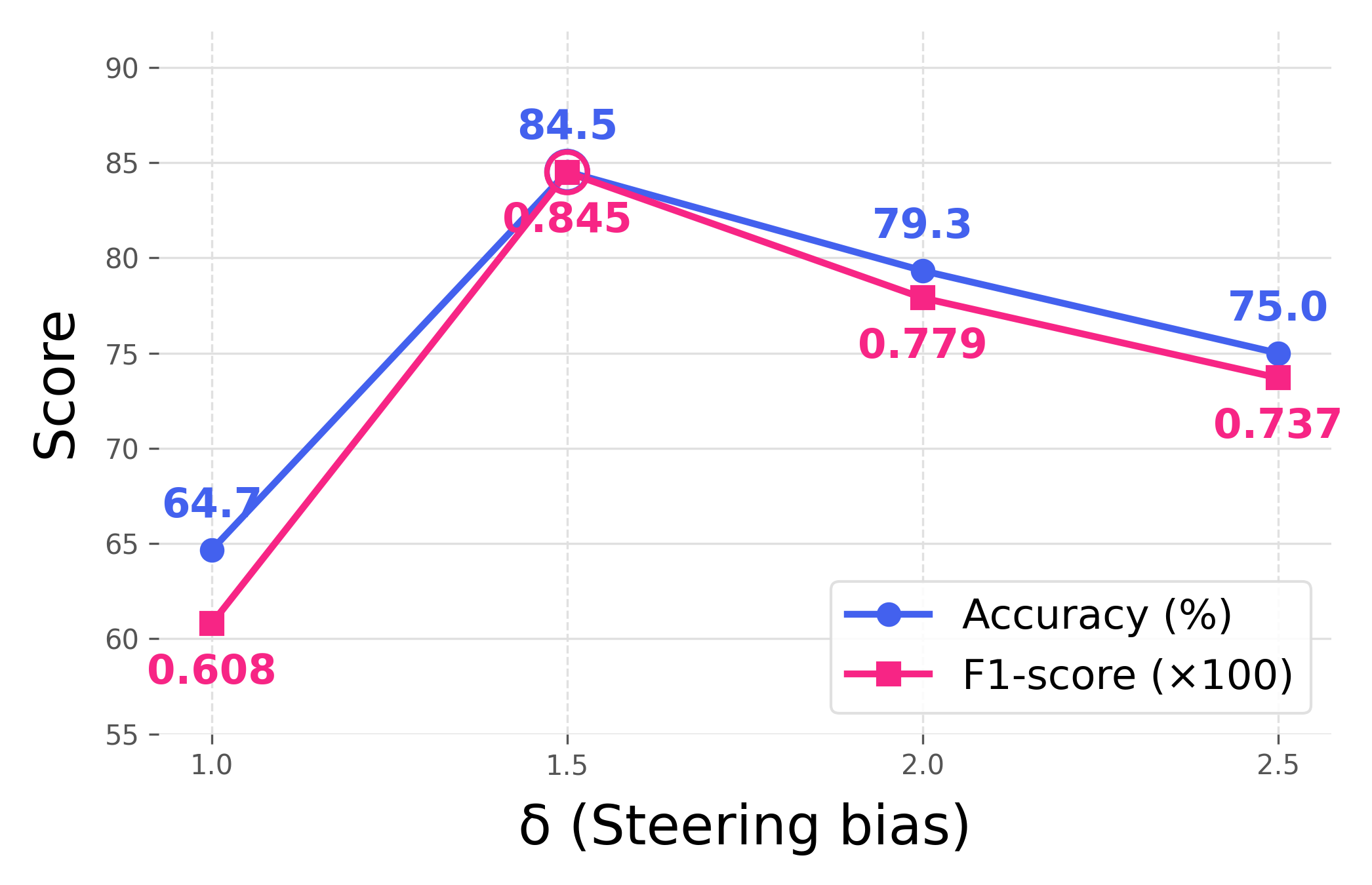}
    \caption{Effect of the steering bias $\delta$ on OSE (Llama3.1 8B). Accuracy and F1 $(\times 100)$ peak at $\delta=1.5$, showing a clear trade-off between steering strength and generation quality.}
    \label{fig:delta}
\end{figure}

\begin{figure*}[h!]
    \centering
    \includegraphics[width=0.85\textwidth]{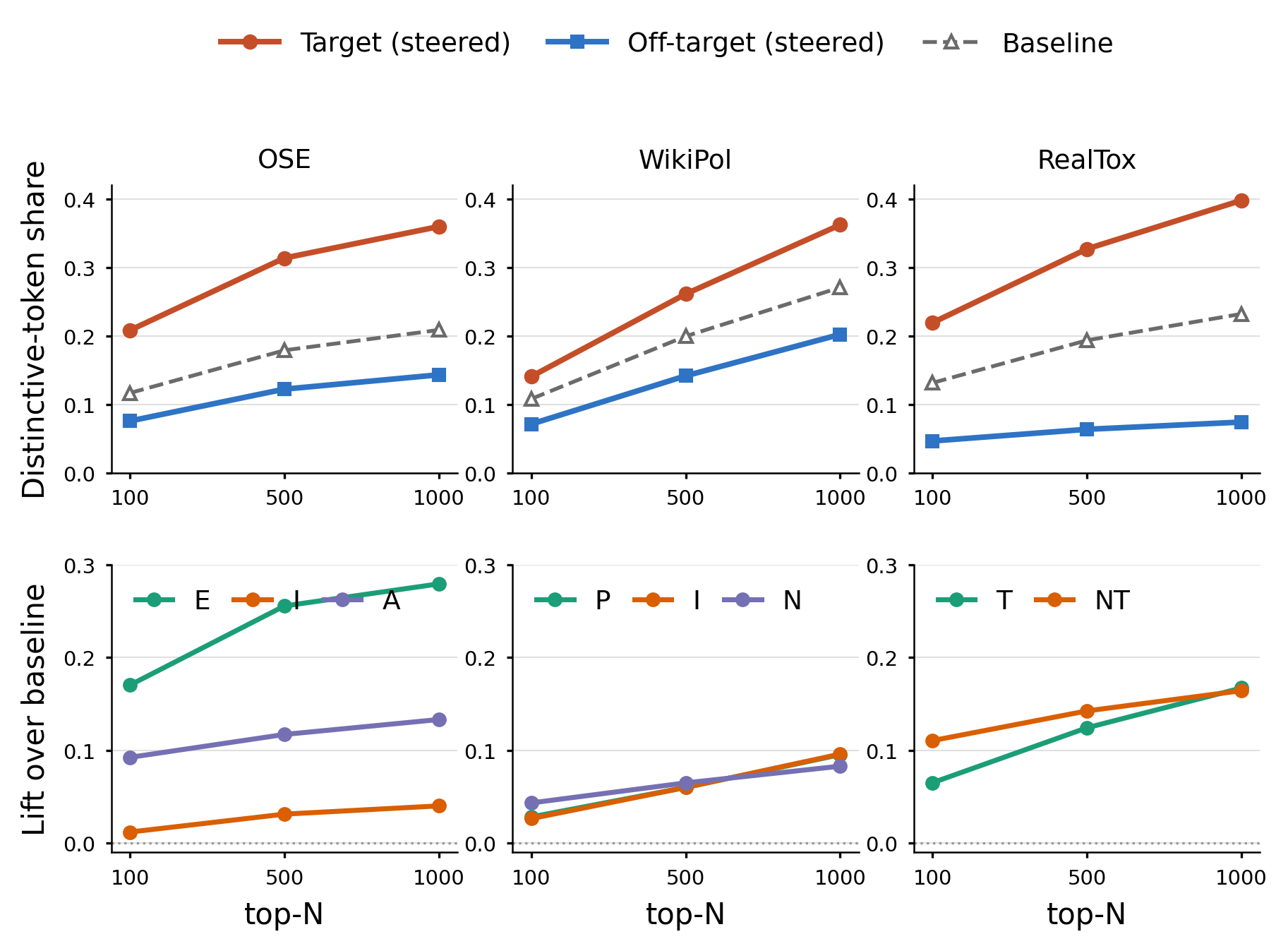}
    \caption{Distinctive-token share of each class in generated text, as a function of set size $N$. Across all $N$, \texttt{SWAI} raises the target-class share (Target) above Baseline while pushing the others (Off-target) below it.}
    \label{fig:selectivity}
\end{figure*}

\section{Analysis}
\subsection{Sensitivity Analysis of Steering Bias}

The steering bias $\delta$ directly controls how strongly the statistical signal
overrides the base model's contextual preferences, making it the central
hyperparameter of \texttt{SWAI}. To examine its effect, we vary $\delta \in \{1.0, 1.5,
2.0, 2.5\}$ on \textsc{OSE} with Llama3.1 8B while holding the candidate size $K$ and the
favored ratio $\rho$ fixed.

As shown in Figure~\ref{fig:delta}, the performance follows a clear inverted-U
pattern, peaking at $\delta=1.5$ with an Accuracy of 84.53\% and an F1 of 0.845,
and degrading monotonically on both sides. When $\delta$ is too small
$(\delta=1.0)$, the additive bias is insufficient to shift the decoding
distribution against the base model's strong contextual prior, and the
statistical score table exerts only marginal influence on token selection (Acc
64.67\%, F1 0.608). Conversely, when $\delta$ is too large $(\delta \in \{2.0,
2.5\})$, the bias begins to dominate the original logits within the candidate
set $C_t$, pushing the sampler toward statistically characteristic but
contextually less plausible tokens. This over-steering erodes local fluency and
long-range coherence, which in turn blurs the target characteristic from the
judge's perspective and lowers both Accuracy and F1.

These results support a key design intuition of the method, since the intervention
is confined to a top-$K$ candidate set already deemed plausible by the base
model, moderate biases are sufficient to realize the target characteristic
without sacrificing generation quality. The robustness of the peak region
(gradual degradation rather than sharp collapse on either side) further
indicates that $\delta$ does not require delicate per-task tuning, and we adopt
$\delta=1.5$ as the default across all experiments.

\begin{figure*}[h!]
    \centering
    \includegraphics[width=0.85\textwidth]{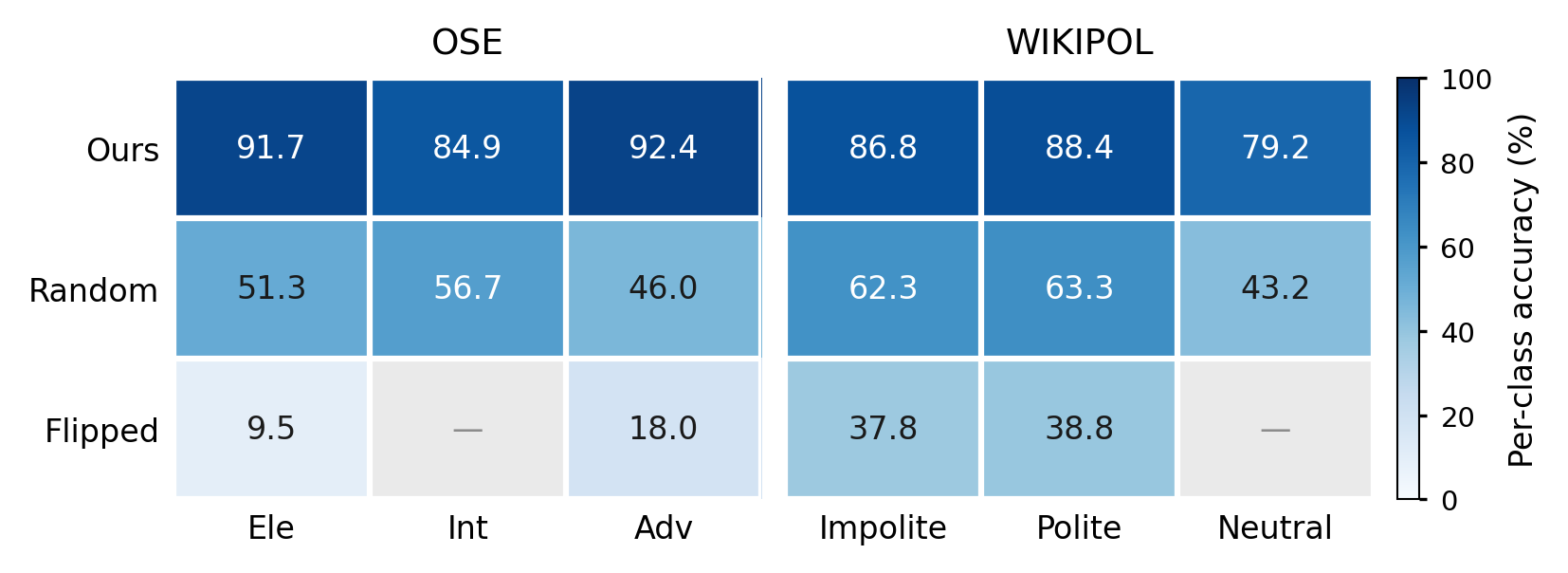}
    \caption{Lookup-table ablation on \textsc{OSE} and \textsc{WikiPol}. Each cell reports
    per-class one-vs-rest accuracy. \textbf{Random} shuffles the score values
    across token IDs, preserving the score distribution while breaking the
    token-to-score mapping; \textbf{Flipped} steers with another class's table and
    is run on one class pair per dataset (\textsc{OSE}: E and A; \textsc{WIKIPOL}: I and P), so the remaining class is left blank.}
    \label{fig:lookup-ablation}
\end{figure*}

\subsection{Selectivity}
We examine whether \texttt{SWAI} intervenes selectively, increasing vocabulary
associated with the target class without also amplifying vocabulary associated
with other classes. For each class, we define its distinctive-token set as the
top-$N$ tokens by token score. For each generated text, we measure the fraction
of tokens that fall within each class-specific set and average this value by
condition. We call this value the distinctive-token share. Unlike judge-based
classification, this metric directly probes steering at the lexical level.
Baseline is the distinctive-token share measured from non-steered generations.
We vary $N$ over $\{100, 500, 1000\}$ to test sensitivity to the set size.

Figure~\ref{fig:selectivity} shows the results across the three datasets. For
all datasets and all values of $N$, \texttt{SWAI} increases the distinctive-token share
of the target class relative to Baseline (top row), while the off-target shares
stay at or below Baseline. Although larger $N$ raises the absolute shares by
expanding each distinctive-token set, this ordering is preserved across all
cutoffs. The per-class lift over Baseline (bottom row) confirms that this gain
holds class by class, though its magnitude varies across classes.

The target-share increase is expected because \texttt{SWAI} uses target-class
token scores during decoding. The key result is that off-target shares are
suppressed rather than jointly increased. This off-diagonal suppression
indicates that \texttt{SWAI} performs directional steering rather than indiscriminate
vocabulary amplification. Further analyses are provided in Appendix~\ref{app:selectivity}.


\subsection{Lookup-Table Ablation}
\label{sec:lookup_ablation}
Our method combines two components, and separating their contributions
clarifies where the steering effect originates. The first is the $\delta$-bias
mechanism, which adds a constant offset to a high-scoring subset of candidate
tokens at each decoding step. The second is the score table, which determines
which tokens receive that offset. If the mechanism alone were sufficient, then
applying it through an uninformative or incorrect table should preserve much of
the control. We test this by corrupting only the table while holding the
mechanism, decoding configuration, random seed, and sample budget fixed, so
that any change in accuracy is attributable to the table rather than to the
procedure. We evaluate two corruptions. \emph{Random} shuffles the score
values across token IDs, which preserves the score distribution but removes the
association between a token and its statistical score. \emph{Flipped} steers
toward a target using the score table of a different class, for instance the
Advanced table when the target is Elementary.

\begin{table}[t]
\centering
\small
\begin{tabular}{lccc}
\toprule
Dataset & $z_r(v)<0$ & $z_r(v)=0$ & $z_r(v)>0$ \\
\midrule
OSE     & 0.1--2.4\% & 14.3--19.4\% & 79.8--85.0\% \\
WikiPol & 1.1--2.2\% & 28.3--36.1\% & 62.8--70.6\% \\
RealTox & 0--7.0\%   & 12.6--17.9\% & 80.4--82.1\% \\
\bottomrule
\end{tabular}
\caption{Distribution of statistical score signs among favored
tokens across datasets. Ranges are taken across target classes.}
\label{tab:favored_score_sign}
\end{table}

Figure~\ref{fig:lookup-ablation} shows that both corruptions reduce per-class
accuracy across every class and both datasets. Relative to the intact table,
Random lowers accuracy by 28 to 46 points on \textsc{OSE} and by 25 to 36 points on
\textsc{WikiPol}. Flipped degrades control further, by up to 82 points on \textsc{OSE} and by
roughly 50 points on \textsc{WikiPol}. The sharpest drop appears on the \textsc{OSE} Elementary
class, where accuracy falls from 91.7\% to 9.5\% under Flipped, indicating that
a wrong-class table drives generation in the opposing direction rather than
merely weakening the signal.
These results separate the mechanism from its input. The $\delta$-bias
procedure does not steer on its own; it steers because the table encodes which
tokens characterize each class. A shuffled table carries no such signal, so
control degrades toward the unsteered regime, whereas a wrong-class table
supplies a reversed signal and degrades control further.
We therefore read the effect as a
directional and attribute-specific intervention rather than a generic
perturbation of the vocabulary distribution.

\begin{table*}[h!]
\centering
\begin{tabular}{lcccccc}
\toprule
Judge & Acc. & F1 & Prec. & Recall & $\kappa$ & MCC \\
\midrule
GPT-4o & 84.53\% & 0.845 & 84.53\% & 84.53\% & 0.768 & 0.768 \\
Gemini 3.5 Flash & 85.19\% & 0.852 & 85.19\% & 85.19\% & 0.778 & 0.779 \\
\bottomrule
\end{tabular}
\caption{Cross-LLM evaluation on OSE. Both judge models yield highly comparable performance on \texttt{SWAI}-generated text.}
\label{tab:cross_llm_eval}
\end{table*}

\begin{table}[h!]
\centering
\small
\begin{tabular}{lccc}
\toprule
 & Elementary & Intermediate & Advanced \\
\midrule
Annotator 1 & 88.24 & 76.47 & 76.47 \\
Annotator 2 & 76.47 & 52.94 & 70.59 \\
Annotator 3 & 70.59 & 58.82 & 70.59 \\
\midrule
\textbf{Overall} & \textbf{78.43} & \textbf{62.75} & \textbf{72.55} \\
\bottomrule
\end{tabular}
\caption{Human evaluation accuracy (\%) for \texttt{SWAI}-generated text across the three reading levels in OSE.}
\label{tab:human_eval}
\end{table}

\subsection{Favored-Token Score Distribution}

We further examine whether the favored set $F_t$ may contain
tokens whose statistical scores are negatively associated with
the target characteristic. Across all datasets, most favored
tokens have positive target-association scores, while
negative-score tokens are rare. As shown in
Table~\ref{tab:favored_score_sign}, positive-score tokens account
for 79.8--85.0\% on OSE, 62.8--70.6\% on WikiPol, and
80.4--82.1\% on RealTox. Negative-score tokens account for only
0--2.4\% in most settings, with the largest proportion being
7.0\% for the toxic RealTox target. The remaining zero-score
tokens are absent from the lookup table and therefore provide
no evidence in either direction. Since \texttt{SWAI} selects the
highest-scoring half of the model's top-100 candidates, tokens
positively associated with the target are strongly prioritized,
making opposite-direction boosting uncommon in practice.

\subsection{Evaluator Robustness}

\paragraph{Cross-LLM evaluation.}
We further evaluate the same generations using Gemini 3.5 Flash as an additional LLM judge. The resulting performance is highly consistent with GPT-4o: Gemini achieves an accuracy of 85.19\% and an F1 score of 0.852, compared with 84.53\% and 0.845 for GPT-4o. The two judges also show strong agreement, with 83.33\% label agreement and Cohen's $\kappa=0.750$. Their continuous evaluation scores are highly correlated (Pearson's $r=0.875$).

\paragraph{Human evaluation.}
To complement the GPT-4o evaluation, we conduct a human evaluation on OSE with three independent annotators: one PhD student in English Language and Literature and two graduate students proficient in English. Annotators followed the same evaluation protocol as GPT-4o, including the reading-level definitions and decision criteria (Appendix~\ref{app:judge_prompt}).

As shown in Table~\ref{tab:human_eval}, human annotators consistently recognize the intended reading levels of \texttt{SWAI}-generated text, achieving overall accuracies of 78.43\%, 62.75\%, and 72.55\% for Elementary, Intermediate, and Advanced, respectively. Inter-annotator agreement is also substantial, with Fleiss' $\kappa=0.634$. These results provide independent evidence that \texttt{SWAI}’s steering effect is not specific to GPT-4o, but reflects perceptible and consistent shifts in reading level.

When the three human annotators and the two LLM judges are considered jointly, agreement remains meaningful, with Fleiss' $\kappa=0.529$. Taken together, these results indicate that \texttt{SWAI}'s improvements are not specific to a particular evaluator: the induced reading-level shifts are consistently recognized across both human and LLM judgments.

\begin{rqbox}
    \textbf{Finding.} A logit steering based on corpus statistics improves generation performance without additional model training and activation intervention.
\end{rqbox}

\section{Conclusion}
We introduce \texttt{SWAI}, a training-free decoding-time method that uses corpus-derived token scores to selectively bias logits within a top-K candidate set. Across readability, politeness, and toxicity control, \texttt{SWAI}
improves controllability over prompt-only steering and remains effective without
modifying model parameters or internal activations. Selectivity and lookup-table
ablations further show that the gains depend on target-specific statistical
scores rather than on generic logit perturbation. These findings highlight the potential of lightweight corpus-statistical signals as a practical and broadly applicable mechanism for controllable generation.

\section*{Limitations}
While the proposed approach is designed to be lightweight and training-free, it
still assumes access to a pretrained language model and the computational
resources required for decoding-time generation. In addition, applying the
method to new output characteristics or datasets requires computing token-level statistics
from labeled corpora, which introduces a modest preprocessing step. Finally, our
experiments focus on large language models and standard text generation
settings; exploring the applicability of the method to other modeling paradigms
or generation frameworks is left for future work.

\section*{Ethical Considerations}

The characteristics that motivate \texttt{SWAI} also constitute its principal ethical risk.
Since the intervention adds a positive bias to tokens that statistically characterize the designated target class, the mechanism is directionally symmetric: a score table constructed for a harmful target could reinforce that characteristic just as a score table constructed for a safety-oriented target can help mitigate it. The Flipped ablation in Section~\ref{sec:lookup_ablation} provides empirical support for this symmetry, as steering with an opposing class's score table reverses the direction of control rather than merely weakening it. This risk is amplified by the lightweight nature of the method, since \texttt{SWAI} requires no training, no auxiliary model, and only access to output logits, thereby lowering the barrier to misuse compared with training-based approaches that require labeled data and additional computation. We therefore release \texttt{SWAI} to support reproducibility and research on safety-oriented controllable generation, and we intend the method for mitigating harmful outputs rather than producing them.

\section*{Acknowledgments}
This work was supported by the KISA grant~(RS-2025-02222626) and  the AI Graduate School Program (RS-2020-II201361).

\bibliography{custom}

\appendix

\section{Experimental Details}
\label{app:details}
\subsection{Baselines}
We provide additional details on the baseline methods used for comparison in Section~\ref{sec:experiment}.

\paragraph{Prompting.}
As a prompt-only baseline, we steer generation solely through natural-language
instructions specifying the desired output characteristic. The target
characteristic is provided to the model together with the input text, and
generation proceeds directly from the model’s original decoding distribution. No
model parameters are updated, and no auxiliary models, internal activations, or
decoding-time logit interventions are used.

\paragraph{GeDi.}
\citet{krause-etal-2021-gedi-generative} train a class-conditional language
model that acts as a generative discriminator. At each decoding step, it
contrasts the likelihoods of the partial sequence under a desired control code
and an undesired anti-control code. Using Bayes’ rule, it estimates the
probability that each candidate next token belongs to the desired class and uses
these probabilities to reweight and filter the base model’s decoding
distribution. Thus, GeDi performs inference-time control but requires a
separately trained class-conditional LM and additional forward passes through
this auxiliary model during generation.

\paragraph{DExperts.}
\citet{liu-etal-2021-dexperts} combine the base model with an expert LM trained
on text exhibiting the desired attribute and an anti-expert LM trained on text
exhibiting the undesired attribute. At each decoding step, it compares the
next-token preferences of the two auxiliary models, promoting tokens favored by
the expert while suppressing those favored by the anti-expert, and then combines
this contrastive signal with the base model’s original predictions. To preserve
fluency, the adjustment is restricted to tokens retained by the base model’s
filtering. DExperts therefore provides direct decoding-time control, but
requires separately fine-tuned auxiliary LMs and additional forward passes
through them at every generation step.

\subsection{Datasets}
We describe below the datasets used in our experiments.

\begin{itemize}
    \item \textsc{OSE}~\citep{vajjala2018onestopenglish} is an English learner corpus consisting of parallel news articles rewritten at three reading levels (elementary, intermediate, and advanced), and is used to evaluate readability and difficulty-controlled text generation.
    \item \textsc{WikiPol}~\citep{danescu2013computational} is a dataset of online request utterances annotated with continuous perceived politeness scores, enabling quantitative analysis of politeness variation in request expressions.
    \item \textsc{RealTox}~\citep{gehman2020realtoxicityprompts} is a large-scale dataset of naturally occurring prompts annotated with toxicity scores, designed to assess the propensity of language models to generate toxic content during continuation.
\end{itemize}

\subsection{Hyperparameters}
\label{app:hyperparam}
We set the Dirichlet smoothing scale to $\alpha=0.01$ and use a pooled prior based
on token frequencies aggregated over the entire corpus. During decoding,
\texttt{SWAI} forms a top-$K$ candidate set at each step and applies logit bias
only within this set, with $K=100$. From the candidate set, we select a favored
subset consisting of the top fraction $\rho=0.5$ of tokens with the highest
statistical scores $z_r(v)$ (\textit{i.e.}, favoring the top 50\% of candidates),
and add a constant bias $\delta=1.5$ to their logits. For \textsc{OSE}, we set
the maximum generation lengths to 650, 800, and 1000 tokens for the
\textit{Elementary}, \textit{Intermediate}, and \textit{Advanced} classes,
respectively, matching the average token lengths of the corresponding classes in
the original dataset. For \textsc{WikiPol} and \textsc{RealTox}, we set the
maximum generation length to 50 tokens.

\subsection{Metrics}
\label{app:metrics}
We evaluate the effectiveness of output characteristic control using standard classification and reliability metrics
computed from the labels predicted by the judge model.
Let $y \in \mathcal{Y}$ denote the target output characteristic label and $\hat{y}$ the label predicted by the judge
for a generated sample.

\paragraph{Accuracy}
Accuracy measures the proportion of correctly predicted samples:
\begin{equation*}
\mathrm{Accuracy} = \frac{1}{N} \sum_{i=1}^{N} \mathbb{I}[\hat{y}_i = y_i],
\end{equation*}
where $N$ is the total number of evaluated samples and $\mathbb{I}[\cdot]$ is the indicator function.
While accuracy provides an intuitive measure of overall performance, it can be misleading under class imbalance,
which motivates the use of additional metrics.

\paragraph{Precision, Recall, and F1-score}
For each class, precision and recall are defined as:
\begin{align*}
\mathrm{Precision} &= \frac{\mathrm{TP}}{\mathrm{TP} + \mathrm{FP}}, \\
\mathrm{Recall}    &= \frac{\mathrm{TP}}{\mathrm{TP} + \mathrm{FN}}.
\end{align*}

where $\mathrm{TP}$, $\mathrm{FP}$, and $\mathrm{FN}$ denote the number of true positives,
false positives, and false negatives, respectively.
The $F_1$-score is given by the harmonic mean of precision and recall:
\begin{equation*}
F_1 =
\frac{2 \cdot \mathrm{Precision} \cdot \mathrm{Recall}}
{\mathrm{Precision} + \mathrm{Recall}}.
\end{equation*}

For multi-class settings (\textsc{OSE} and \textsc{WikiPol}), we report macro-$F_1$, which assigns equal weight
to each class and prevents dominant classes from overshadowing minority or ambiguous categories.

\paragraph{Cohen's Kappa ($\kappa$)}
Cohen's Kappa measures the agreement between predicted and target labels while correcting for chance agreement:
\begin{equation*}
\kappa = \frac{p_o - p_e}{1 - p_e},
\end{equation*}
where $p_o$ denotes the observed agreement and $p_e$ the expected agreement under random labeling.
Unlike accuracy, $\kappa$ penalizes degenerate solutions that collapse predictions into a single class,
making it particularly informative for evaluating controllability under skewed label distributions.

\paragraph{Matthews Correlation Coefficient (MCC)}
MCC provides a balanced measure of classification quality
that accounts for all entries of the confusion matrix.
In the multi-class setting, MCC is defined as:
\begin{equation*}
\mathrm{MCC}
=
\frac{TP/N - S P}
{\sqrt{S P (1 - S)(1 - P)}}.
\end{equation*}

where
$N = TP + TN + FP + FN$ is the total number of instances,
$S = (TP + FN)/N$ is the proportion of true positive instances,
and
$P = (TP + FP)/N$ is the proportion of positive predictions.

\paragraph{Confidence}
In addition to label-based metrics, we report the confidence score produced by the judge model,
which reflects the model's certainty in its predicted label.
This measure provides complementary insight into whether improvements in classification performance
correspond to increased decisiveness, or whether a method exhibits overconfident or underconfident behavior.

\section{Characteristics and Advantages}
\label{subsec:advantages}
Our framework offers several technical advantages over existing
output characteristic control methodologies:
\begin{itemize}
    \item \textbf{Architecture-agnostic Logit Steering:} By operating exclusively on output logits,
    \texttt{SWAI} is applicable to any model architecture without requiring access to internal hidden
    states or specialized gradients.
    \item \textbf{Granular Control Interface:} We provide a mechanism for precise navigation
    of the trade-off between control weight and text quality through three primary hyperparameters:
    the bias magnitude ($\delta$), the number of steering targets ($m$), and the candidate set size ($C_t$).
    \item \textbf{Computational Efficiency}: Our logit steering relies solely on table lookups and vector
    additions, ensuring that the computational overhead during inference is negligible compared to fine-tuning
    or activation-based steering.
\end{itemize}

\section{Open-ended Generation}

\begin{table*}[t]
\centering
\begin{tabular}{l|c|cccccc}
\toprule
Model & Level & Accuracy & $F_1$ & Precision & Recall & $\kappa$ & MCC \\
\midrule
\multirow{4}{*}{Llama3.1 8B}
 & \textit{Ele}   & \textbf{88.93\%} & \textbf{0.834} & \textbf{83.40\%} & \textbf{83.40\%} & \multirow{4}{*}{0.672} & \multirow{4}{*}{0.672} \\
 & \textit{Int} & 78.93\% & 0.684 & 68.40\% & 68.40\% &  &  \\
 & \textit{Adv} & 88.40\% & 0.826 & 82.60\% & 82.60\% &  &  \\
 & Total & 78.13\% & 0.781 & 78.13\% & 78.13\% &  &  \\
\midrule
\multirow{4}{*}{Llama3.2 1B}
 & \textit{Ele}   & \textbf{85.56\%} & \textbf{0.783} & \textbf{78.52\%} & \textbf{78.00\%} & \multirow{4}{*}{0.553} & \multirow{4}{*}{0.553} \\
 & \textit{Int} & 72.00\% & 0.583 & 57.89\% & 58.67\% &  &  \\
 & \textit{Adv} & 82.89\% & 0.743 & 74.50\% & 74.00\% &  &  \\
 & Total & 70.22\% & 0.703 & 70.30\% & 70.22\% & &  \\
\bottomrule
\end{tabular}
\caption{Results of open-ended generation on \textsc{OSE}.
The model freely continues the input text and the generated outputs are evaluated
for reading difficulty using the Judge LLM.}
\label{tab:openended}
\end{table*}

For the \textsc{OSE} dataset, we further analyze the effect of logit steering in
an open-ended generation setting beyond the standard paraphrase setup
(Table~\ref{tab:openended}). Specifically, given one or two initial sentences,
the model is instructed to freely continue the text, and the generated outputs
are subsequently evaluated for reading difficulty (E/I/A). As shown by the
class-wise Accuracy and $F_1$ scores in Table~\ref{tab:openended}, we observe
stable class separation with relatively high accuracy across all difficulty
levels, despite the increased generation length and the higher risk of content
drift in this setting.

Examining performance by difficulty level, the \textit{Advanced} class achieves
the highest accuracy for both models, suggesting that features characterizing
higher difficulty, such as advanced vocabulary and more complex syntactic
structures, can be effectively reinforced through logit steering. In contrast,
the \textit{Intermediate} class exhibits relatively lower Accuracy and $F_1$, which can
be attributed to its inherently ambiguous position between \textit{Elementary}
and \textit{Advanced} in the \textsc{OSE} dataset. This pattern is further
corroborated by the fact that the judge's confidence scores are lowest for the
\textit{Intermediate} class, indicating that the observed errors are more likely due to
overlapping difficulty definitions than instability in the steering mechanism.

Across model scales, Llama3.1 8B consistently outperforms Llama3.2 1B on accuracy and reliability, 
underscoring the role of sufficient model capacity in sustaining steering effects 
for open-ended generation requiring long-range coherence and stylistic consistency.

Overall, the results in Table~\ref{tab:openended} demonstrate that the proposed
logit steering approach enables persistent control over target difficulty levels
in unconstrained, long-form generation, without additional training or
fine-tuning, positioning it as a practical alternative between prompt-based and
training-based control methods.

\section{Per-condition Selectivity}
\label{app:selectivity}

\begin{figure}[h!]
    \centering
    \includegraphics[width=\linewidth]{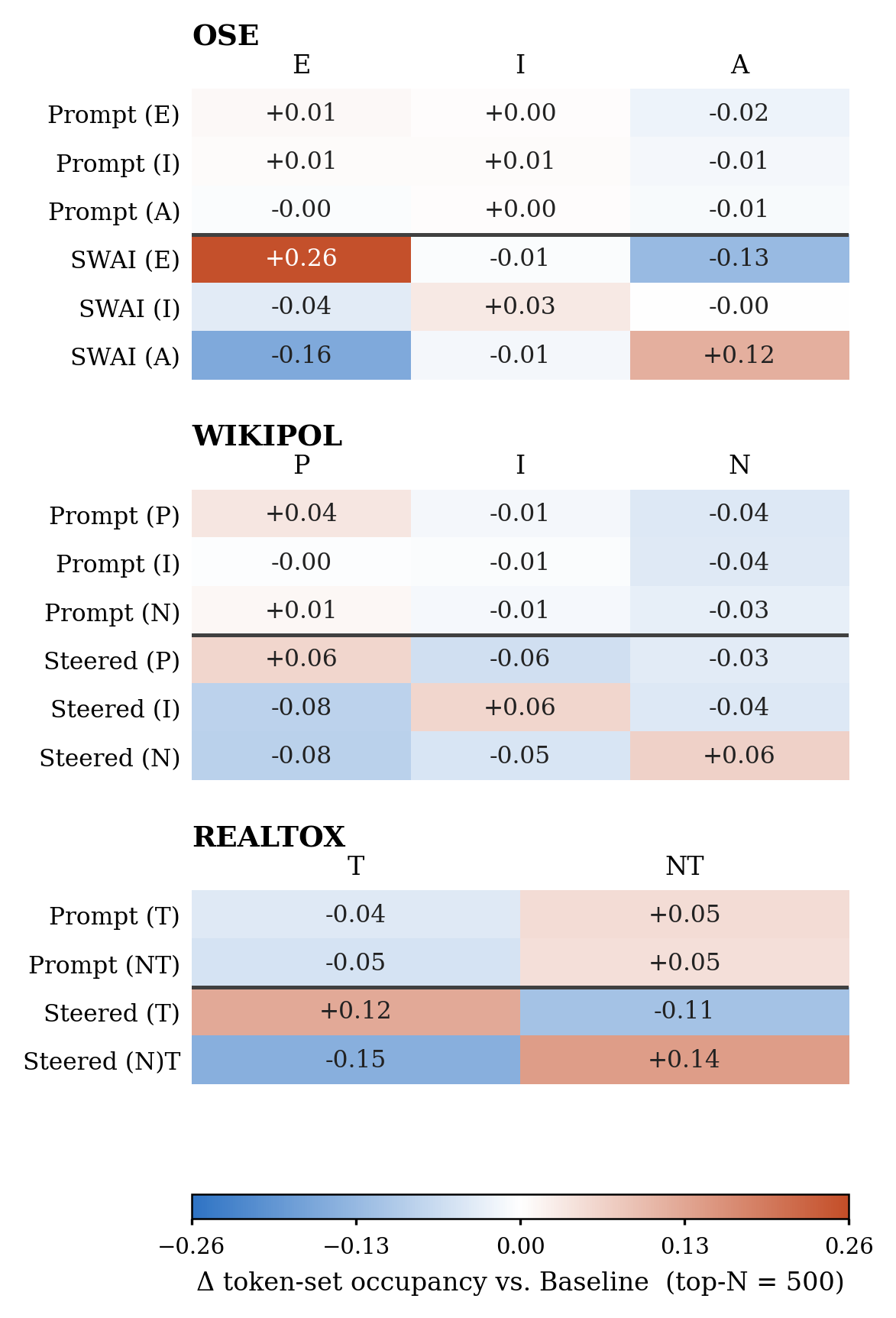}
    \caption{Per-condition selectivity. Change in each class's distinctive-token share relative to Baseline ($\Delta$, top-$N$=500). Steered rows show positive on-diagonal and negative off-diagonal entries, whereas prompt-only rows are near zero.}
    \label{fig:per-class-selectivity}
\end{figure}

We further decompose the Selectivity analysis in the main text by generation condition and measured class, and additionally include the prompt-only condition. Each cell in Figure~\ref{fig:per-class-selectivity} reports the change in distinctive-token share relative to Baseline, defined as $\Delta = \mathrm{share} - \mathrm{Baseline}$, for texts generated under the corresponding condition. We use $N=500$ as the representative set size; robustness to the cutoff is established in the main text.

Rows indicate the generation condition, including prompt-only and steered generations, while columns indicate the class whose distinctive-token share is being measured. Diagonal cells correspond to on-target effects, and off-diagonal cells correspond to off-target effects. Positive values indicate that the corresponding class-specific vocabulary is amplified relative to Baseline, whereas negative values indicate that it is suppressed.

For steered generations, the diagonal cells are positive and the off-diagonal cells are negative. Thus, the directional pattern shown in the main text through averaged curves holds consistently at the level of individual condition--class cells.

By contrast, the prompt-only rows remain close to zero across all cells ($|\Delta| < 0.05$). This indicates that explicit prompting produces little change in the lexical distribution.

\section{Case Study}
\label{app:case_study}
The quantitative results show that \texttt{SWAI} shifts generated text toward the target reading level, but they do not reveal where the intervention affects token selection. To inspect this process, we instrument the decoder at each generation step and record whether the sampled token belonged to the favored set $F_t$ and whether the bias $\delta$ changed the top-ranked candidate.

\begin{figure*}[t]
  \centering
  \begin{subfigure}{\textwidth}
    \centering
    \includegraphics[width=0.9\textwidth]{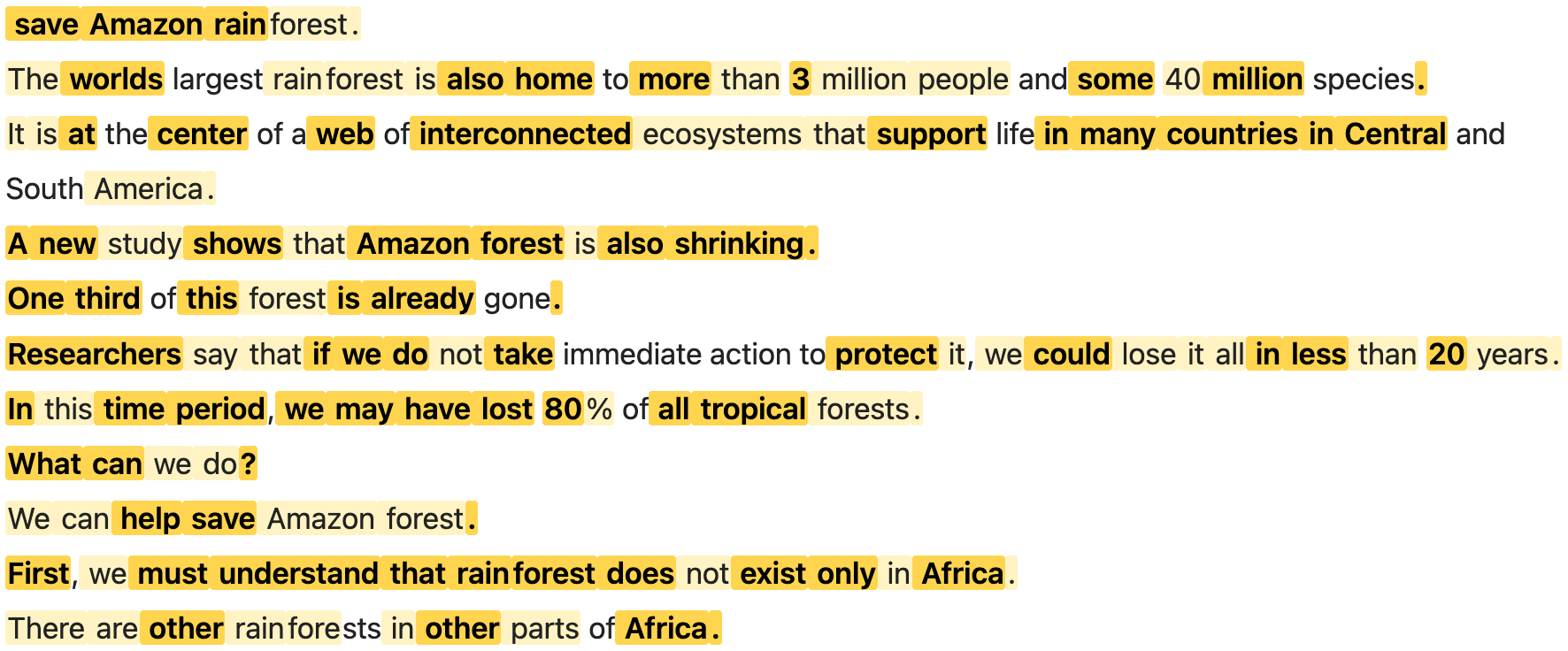}
    \caption{Elementary}
    \label{fig:case_e}
  \end{subfigure}
  \par\medskip
  \begin{subfigure}{\textwidth}
    \centering
    \includegraphics[width=0.9\textwidth]{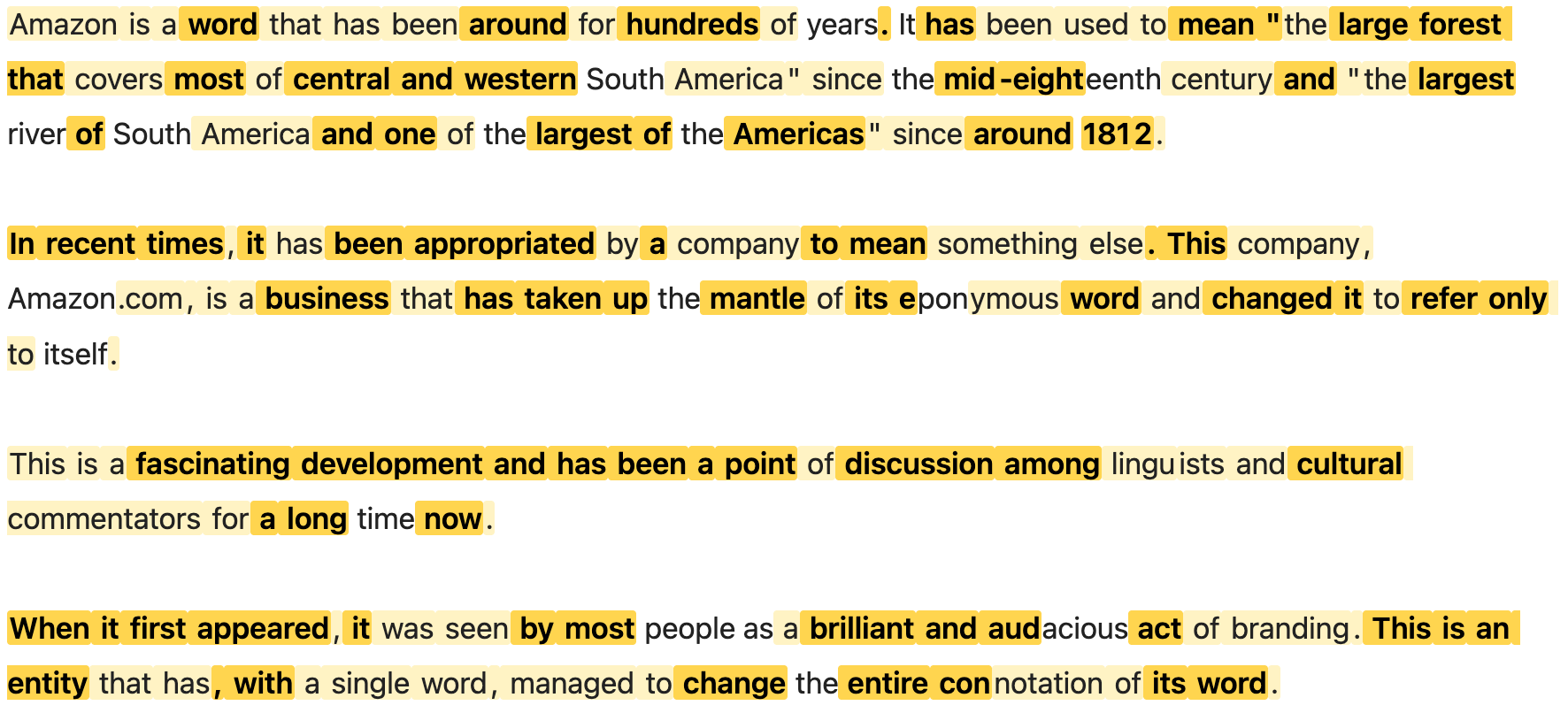}
    \caption{Intermediate}
    \label{fig:case_i}
  \end{subfigure}
  \par\medskip
  \begin{subfigure}{\textwidth}
    \centering
    \includegraphics[width=0.9\textwidth]{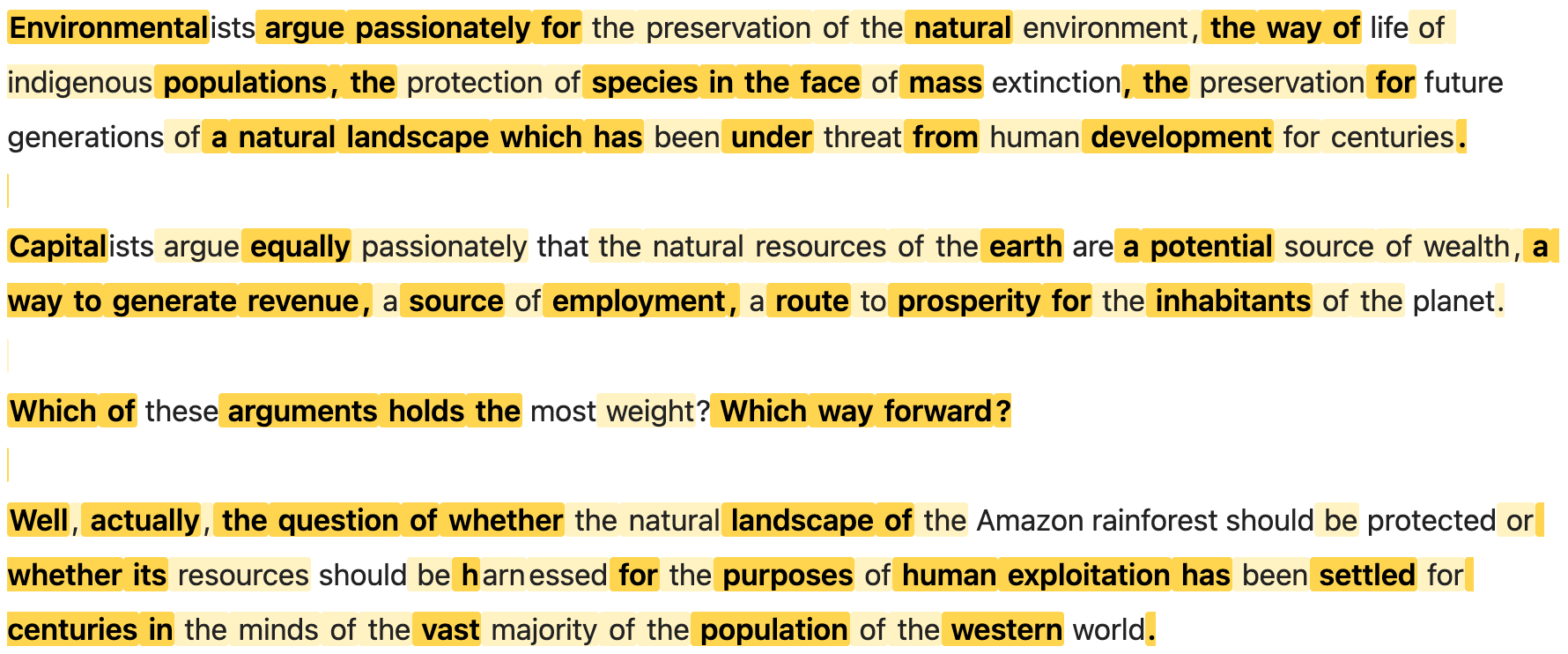}
    \caption{Advanced}
    \label{fig:case_a}
  \end{subfigure}
  \caption{Qualitative promoted-token examples for \textsc{OSE} on the shared topic of `Amazon'. Dark highlights indicate tokens promoted by the logit bias; light highlights indicate biased tokens that were already top-ranked before intervention.}
  \label{fig:case_study}
\end{figure*}

Figure~\ref{fig:case_study} visualizes one representative OSE generation for each reading level. Dark highlights mark \emph{decisive} promotions: tokens that belonged to $F_t$ and became the top-ranked candidate only after the bias was applied. Light highlights mark \emph{concordant} tokens: tokens that belonged to $F_t$ but were already top-ranked under the original logits $z_t$. Unhighlighted tokens were not biased, either because they fell outside the candidate set $C_t$ or below the favored threshold.

The highlighted tokens reveal level-specific steering behavior. In the Elementary example, decisive promotions concentrate on simple framing and repetitive lexical choices, including high-frequency words and repeated content words such as \textit{forest}, \textit{other}, and \textit{also}. They also support short, plain evaluative phrasing, as in \textit{Logging is not good}. These promotions align with the simplification signals associated with learner-oriented prose. In the Advanced example, decisive promotions instead favor abstract and information-dense vocabulary, including \textit{populations}, \textit{exploitation}, \textit{prosperity}, \textit{revenue}, and \textit{inhabitants}, while sustaining a parallel argumentative contrast between \textit{Environmentalists} and \textit{Capitalists}. The Intermediate example falls between these patterns, combining accessible narration with occasional more sophisticated promotions such as \textit{appropriated}, \textit{audacious}, and \textit{mantle}.

These examples support two observations. First, decisive promotions occur throughout the generations, indicating that \texttt{SWAI} actively changes token selection rather than merely labeling tokens the base model would already have chosen. Second, because $F_t$ is selected within the top-$K$ candidate set, promoted tokens remain among the base model's plausible continuations. \texttt{SWAI} therefore redirects choices among plausible tokens instead of forcing arbitrary substitutions. Concordant tokens further show that the statistical signal sometimes reinforces the base model's own preference.

The examples in Figure~\ref{fig:case_study} are selected for readability and are intended as qualitative illustrations rather than quantitative evidence.

\begin{table*}[t]
\centering
\footnotesize
\setlength{\tabcolsep}{5pt}
\begin{tabular}{r l l @{\hspace{2.2em}} r l l}
\toprule
\textbf{Rank} & \textbf{T} & \textbf{NT} & \textbf{Rank} & \textbf{T} & \textbf{NT} \\
\midrule
1 & \tk{\bd{}you}~(39.69) & \tk{\bd{}\bd{}}~(38.40) & 26 & \tk{\bd{}idiot}~(16.86) & \tk{\bd{}state}~(9.87) \\
2 & \tk{\bd{}shit}~(34.99) & \tk{\bd{}the}~(28.48) & 27 & \tk{\bd{}breasts}~(16.22) & \tk{200}~(9.86) \\
3 & \tk{\bd{}\textquotedblleft{}}~(34.68) & \tk{.}~(22.60) & 28 & \tk{\bd{}raped}~(16.02) & \tk{3}~(9.48) \\
4 & \tk{\bd{}sex}~(31.20) & \tk{\bd{}in}~(22.14) & 29 & \tk{\bd{}rape}~(15.89) & \tk{\bd{}season}~(9.38) \\
5 & \tk{\bd{}stupid}~(30.47) & \tk{201}~(18.74) & 30 & \tk{\bd{}suck}~(15.72) & \tk{9}~(9.32) \\
6 & \tk{\bd{}her}~(29.19) & \tk{\bd{}its}~(14.47) & 31 & \tk{\bd{}crap}~(15.63) & \tk{.S}~(9.26) \\
7 & \tk{\bd{}I}~(28.48) & \tk{\bd{}of}~(13.81) & 32 & \tk{\bd{}\textquoteleft{}}~(15.62) & \tk{2}~(9.06) \\
8 & \tk{\bd{}penis}~(27.89) & \tk{The}~(13.19) & 33 & \tk{\bd{}black}~(15.40) & \tk{\bd{}year}~(8.88) \\
9 & \tk{\bd{}ass}~(27.87) & \tk{\bd{}from}~(13.00) & 34 & \tk{\bd{}genitals}~(15.40) & \tk{\bd{}U}~(8.75) \\
10 & \tk{\bd{}fucking}~(27.38) & \tk{\bd{}new}~(12.59) & 35 & \tk{\bd{}woman}~(15.32) & \tk{\bd{}company}~(8.67) \\
11 & \tk{\bd{}fuck}~(24.38) & \tk{\bd{}has}~(12.46) & 36 & \tk{\bd{}she}~(15.25) & \tk{\bd{}announced}~(8.64) \\
12 & \tk{\bd{}"}~(23.66) & \tk{1}~(12.43) & 37 & \tk{\bd{}dumb}~(15.24) & \tk{\bd{}according}~(8.44) \\
13 & \tk{.\textquotedblright{}}~(23.37) & \tk{\bd{}\$}~(12.35) & 38 & \tk{!\textquotedblright{}}~(15.03) & \tk{\bd{}by}~(8.38) \\
14 & \tk{\bd{}kill}~(21.96) & \tk{7}~(11.88) & 39 & \tk{***}~(14.94) & \tk{\bd{}since}~(8.37) \\
15 & \tk{\bd{}like}~(21.30) & \tk{4}~(11.80) & 40 & \tk{,}~(14.92) & \tk{\bd{}system}~(7.98) \\
16 & \tk{I}~(20.58) & \tk{5}~(11.75) & 41 & \tk{\textquoteright{}re}~(14.82) & \tk{\bd{}first}~(7.86) \\
17 & \tk{\bd{}f}~(20.53) & \tk{\bd{}government}~(11.35) & 42 & \tk{."}~(14.79) & \tk{\bd{}team}~(7.78) \\
18 & \tk{\bd{}your}~(20.08) & \tk{\bd{}will}~(11.11) & 43 & \tk{\bd{}don}~(14.68) & \tk{\bd{}available}~(7.74) \\
19 & \tk{\bd{}my}~(20.03) & \tk{\bd{}at}~(10.49) & 44 & \tk{\bd{}women}~(14.52) & \tk{\bd{}on}~(7.74) \\
20 & \tk{\bd{}bullshit}~(19.49) & \tk{000}~(10.43) & 45 & \tk{\textquoteright{}m}~(14.31) & \tk{\bd{}data}~(7.63) \\
21 & \tk{\bd{}me}~(19.01) & \tk{0}~(10.43) & 46 & \tk{\bd{}man}~(14.25) & \tk{\bd{}information}~(7.62) \\
22 & \tk{\bd{}him}~(18.69) & \tk{6}~(10.31) & 47 & \tk{**}~(14.24) & \tk{\bd{}officials}~(7.61) \\
23 & \tk{\bd{}bitch}~(18.49) & \tk{\bd{}million}~(10.28) & 48 & \tk{\bd{}then}~(14.13) & \tk{\bd{}percent}~(7.61) \\
24 & \tk{\bd{}damn}~(17.16) & \tk{\bd{}for}~(10.23) & 49 & \tk{\bd{}'}~(14.11) & \tk{\bd{}support}~(7.60) \\
25 & \tk{\bd{}vagina}~(17.10) & \tk{8}~(10.18) & 50 & \tk{\textquoteright{}t}~(14.09) & \tk{\bd{}US}~(7.58) \\
\bottomrule
\end{tabular}
\caption{Top-50 tokens by statistical score for \textsc{RealTox}. \textbf{T}: toxic; \textbf{NT}: non-toxic. The leading dot (\bd{}) marks a token-internal word boundary. Scores are z-normalized one-vs-rest log-odds.}
\label{tab:facevalid-realtox}
\end{table*}
 
\begin{table*}[t]
\centering
\scriptsize
\setlength{\tabcolsep}{2.5pt}
\begin{tabular}{r l l l @{\hspace{0.5em}} r l l l}
\toprule
\textbf{Rank} & \textbf{P} & \textbf{I} & \textbf{N} & \textbf{Rank} & \textbf{P} & \textbf{I} & \textbf{N} \\
\midrule
1 & \tk{Thanks}~(11.89) & \tk{\bd{}why}~(10.07) & \tk{\bd{}\textless{}}~(5.08) & 26 & \tk{\bd{}happy}~(3.64) & \tk{And}~(3.50) & \tk{\bd{}one}~(2.38) \\
2 & \tk{\bd{}help}~(10.29) & \tk{\bd{}Why}~(8.59) & \tk{url}~(4.71) & 27 & \tk{.}~(3.59) & \tk{How}~(3.48) & \tk{\bd{}Do}~(2.36) \\
3 & \tk{\bd{}for}~(9.31) & \tk{\textquoteright{}t}~(7.29) & \tk{\bd{}has}~(4.21) & 28 & \tk{\bd{}OK}~(3.57) & \tk{\bd{}exactly}~(3.45) & \tk{\bd{}(}~(2.35) \\
4 & \tk{\bd{}please}~(8.69) & \tk{Why}~(7.11) & \tk{\textgreater{},}~(3.91) & 29 & \tk{\bd{}By}~(3.52) & \tk{\bd{}all}~(3.45) & \tk{\bd{}where}~(2.34) \\
5 & \tk{\bd{}you}~(8.20) & \tk{\bd{}not}~(6.09) & \tk{\textgreater{}}~(3.59) & 30 & \tk{\bd{}good}~(3.51) & \tk{\bd{}off}~(3.42) & \tk{I}~(2.30) \\
6 & \tk{\bd{}Would}~(7.53) & \tk{\bd{}And}~(5.52) & \tk{\bd{}changed}~(3.49) & 31 & \tk{\bd{}me}~(3.51) & \tk{\bd{}"}~(3.40) & \tk{\bd{}official}~(2.28) \\
7 & \tk{Hi}~(7.52) & \tk{\bd{}did}~(5.42) & \tk{\bd{}they}~(3.42) & 32 & \tk{Hey}~(3.49) & \tk{\bd{}come}~(3.30) & \tk{\bd{}vote}~(2.28) \\
8 & \tk{\bd{}could}~(7.28) & \tk{\textquoteright{}\textquoteright{}}~(5.13) & \tk{\bd{}there}~(3.12) & 33 & \tk{\bd{}:)}~(3.49) & \tk{?}~(3.25) & \tk{\bd{}or}~(2.26) \\
9 & \tk{!}~(7.23) & \tk{You}~(5.07) & \tk{\bd{}sort}~(2.92) & 34 & \tk{\bd{}Can}~(3.44) & \tk{\bd{}evidence}~(3.23) & \tk{person}~(2.25) \\
10 & \tk{Thank}~(6.95) & \tk{\textquoteright{}s}~(4.79) & \tk{\bd{}their}~(2.90) & 35 & \tk{\bd{}more}~(3.43) & \tk{\textquoteright{}\textquoteright{}\textquoteright{}}~(3.21) & \tk{\bd{}show}~(2.24) \\
11 & \tk{\bd{}I}~(6.73) & \tk{\bd{}\textquoteright{}\textquoteright{}}~(4.69) & \tk{\bd{}name}~(2.73) & 36 & \tk{\bd{}your}~(3.42) & \tk{\bd{}yourself}~(3.21) & \tk{\bd{}old}~(2.24) \\
12 & \tk{\bd{}thanks}~(6.04) & \tk{,"}~(4.55) & \tk{\bd{}until}~(2.70) & 37 & \tk{Nice}~(3.38) & \tk{\bd{}revert}~(3.21) & \tk{Is}~(2.23) \\
13 & \tk{\bd{}Could}~(5.81) & \tk{\bd{}really}~(4.55) & \tk{\bd{}Which}~(2.69) & 38 & \tk{\bd{}like}~(3.32) & \tk{\bd{}allowed}~(3.20) & \tk{\bd{}Or}~(2.22) \\
14 & \tk{\bd{}interested}~(5.69) & \tk{"?}~(4.39) & \tk{\bd{}recent}~(2.67) & 39 & \tk{\textquoteright{}ve}~(3.26) & \tk{Are}~(3.18) & \tk{\bd{}thought}~(2.22) \\
15 & \tk{\bd{}great}~(4.88) & \tk{\bd{}who}~(4.18) & \tk{2}~(2.63) & 40 & \tk{\bd{}:-)}~(3.24) & \tk{ly}~(3.18) & \tk{=}~(2.18) \\
16 & \tk{\bd{}willing}~(4.70) & \tk{???}~(4.01) & \tk{\bd{}box}~(2.58) & 41 & \tk{\bd{}map}~(3.24) & \tk{\bd{}player}~(3.15) & \tk{\bd{}expert}~(2.18) \\
17 & \tk{\bd{}mind}~(4.63) & \tk{\bd{}remove}~(3.97) & \tk{The}~(2.58) & 42 & \tk{\bd{}ok}~(3.23) & \tk{\bd{}calling}~(3.07) & \tk{en}~(2.17) \\
18 & \tk{Good}~(4.44) & \tk{??}~(3.84) & \tk{\bd{}wik}~(2.54) & 43 & \tk{\bd{}advice}~(3.20) & \tk{\bd{}attack}~(3.05) & \tk{\bd{}characters}~(2.17) \\
19 & \tk{\bd{}sorry}~(4.31) & \tk{\bd{}answer}~(3.82) & \tk{\bd{}Is}~(2.51) & 44 & \tk{\bd{}sub}~(3.20) & \tk{\bd{}stop}~(3.05) & \tk{Did}~(2.17) \\
20 & \tk{\textquoteright{}d}~(3.91) & \tk{\bd{}people}~(3.78) & \tk{\bd{}author}~(2.49) & 45 & \tk{Hello}~(3.18) & \tk{\bd{}personal}~(3.05) & \tk{\bd{}rationale}~(2.17) \\
21 & \tk{\bd{}work}~(3.91) & \tk{\bd{}You}~(3.71) & \tk{\bd{}five}~(2.47) & 46 & \tk{\bd{}stuff}~(3.14) & \tk{\bd{}what}~(3.02) & \tk{1}~(2.16) \\
22 & \tk{\bd{}reply}~(3.84) & \tk{\bd{}edit}~(3.69) & \tk{\bd{}sourced}~(2.47) & 47 & \tk{\bd{}on}~(3.13) & \tk{He}~(2.99) & \tk{\bd{}true}~(2.14) \\
23 & \tk{\bd{}able}~(3.74) & \tk{\bd{}talking}~(3.66) & \tk{\bd{}example}~(2.43) & 48 & \tk{\bd{}maps}~(3.12) & \tk{\bd{}T}~(2.95) & \tk{\bd{}both}~(2.13) \\
24 & \tk{\bd{}if}~(3.73) & \tk{\textquoteright{}re}~(3.61) & \tk{\bd{}mean}~(2.40) & 49 & \tk{\bd{}friend}~(3.11) & \tk{you}~(2.95) & \tk{It}~(2.09) \\
25 & \tk{\bd{}thoughts}~(3.64) & \tk{So}~(3.59) & \tk{\bd{}protected}~(2.38) & 50 & \tk{\bd{}finished}~(3.11) & \tk{\bd{}let}~(2.94) & \tk{\bd{}the}~(2.09) \\
\bottomrule
\end{tabular}
\caption{Top-50 tokens by statistical score for \textsc{WikiPol}. \textbf{P}: polite; \textbf{I}: impolite; \textbf{N}: neutral. The leading dot (\bd{}) marks a token-internal word boundary.}
\label{tab:facevalid-wikipol}
\end{table*}
 
\begin{table*}[t]
\centering
\scriptsize
\setlength{\tabcolsep}{2.5pt}
\begin{tabular}{r l l l @{\hspace{0.5em}} r l l l}
\toprule
\textbf{Rank} & \textbf{E} & \textbf{I} & \textbf{A} & \textbf{Rank} & \textbf{E} & \textbf{I} & \textbf{A} \\
\midrule
1 & \tk{\bd{}people}~(13.74) & \tk{\bd{}However}~(3.02) & \tk{\bd{}being}~(6.86) & 26 & \tk{\bd{}It}~(4.77) & \tk{\bd{}due}~(1.60) & \tk{\bd{}across}~(3.34) \\
2 & \tk{.}~(12.93) & \tk{m}~(2.87) & \tk{\bd{}as}~(6.65) & 27 & \tk{\bd{}good}~(4.75) & \tk{\bd{}celebrate}~(1.58) & \tk{\bd{}set}~(3.28) \\
3 & \tk{\bd{}very}~(8.46) & \tk{bn}~(2.62) & \tk{\bd{}by}~(6.12) & 28 & \tk{\bd{}started}~(4.70) & \tk{\bd{}confidential}~(1.58) & \tk{\bd{}having}~(3.26) \\
4 & \tk{\bd{}they}~(8.36) & \tk{\bd{}mainly}~(2.11) & \tk{\bd{}out}~(5.76) & 29 & \tk{\bd{}big}~(4.69) & \tk{\bd{}been}~(1.57) & \tk{\bd{}to}~(3.25) \\
5 & \tk{\bd{}million}~(7.96) & \tk{\bd{}charge}~(1.98) & \tk{,}~(5.60) & 30 & \tk{\bd{}give}~(4.60) & \tk{\bd{}chosen}~(1.55) & \tk{\bd{}those}~(3.24) \\
6 & \tk{\bd{}says}~(7.60) & \tk{\bd{}had}~(1.96) & \tk{ed}~(5.44) & 31 & \tk{\bd{}lot}~(4.56) & \tk{\bd{}corner}~(1.54) & \tk{\bd{}into}~(3.15) \\
7 & \tk{\bd{}will}~(7.33) & \tk{\bd{}demanding}~(1.95) & \tk{\bd{}of}~(5.24) & 32 & \tk{\bd{}do}~(4.55) & \tk{-breaking}~(1.54) & \tk{\bd{}largely}~(3.14) \\
8 & \tk{\bd{}They}~(6.95) & \tk{\bd{}provides}~(1.92) & \tk{ing}~(4.84) & 33 & \tk{\bd{}dangerous}~(4.46) & \tk{According}~(1.54) & \tk{\bd{}tend}~(3.13) \\
9 & \tk{\bd{}billion}~(6.93) & \tk{\bd{}greater}~(1.91) & \tk{\bd{}off}~(4.59) & 34 & \tk{\bd{}important}~(4.29) & \tk{\bd{}preferences}~(1.51) & \tk{\bd{}following}~(3.08) \\
10 & \tk{\bd{}money}~(6.61) & \tk{\bd{}ability}~(1.89) & \tk{\bd{}up}~(4.54) & 35 & \tk{\bd{}buy}~(4.17) & \tk{\bd{}resist}~(1.51) & \tk{ating}~(3.07) \\
11 & \tk{\bd{}because}~(6.52) & \tk{\bd{}formation}~(1.86) & \tk{\bd{}which}~(4.53) & 36 & \tk{\bd{}The}~(4.13) & \tk{\bd{}enforce}~(1.51) & \tk{\bd{}proposed}~(3.07) \\
12 & \tk{\bd{}But}~(6.34) & \tk{\bd{}increasing}~(1.83) & \tk{ly}~(4.45) & 37 & \tk{\bd{}we}~(4.11) & \tk{-edge}~(1.51) & \tk{\bd{}keen}~(3.07) \\
13 & \tk{\bd{}want}~(6.25) & \tk{\bd{}expected}~(1.78) & \tk{led}~(4.38) & 38 & \tk{\bd{}there}~(4.08) & \tk{-den}~(1.51) & \tk{\bd{}further}~(3.03) \\
14 & \tk{\bd{}use}~(6.23) & \tk{\bd{}conditions}~(1.77) & \tk{\bd{}while}~(4.35) & 39 & \tk{\bd{}say}~(4.05) & \tk{\bd{}deal}~(1.50) & \tk{\bd{}scale}~(2.99) \\
15 & \tk{\bd{}He}~(5.90) & \tk{\bd{}by}~(1.75) & \tk{\bd{}with}~(4.31) & 40 & \tk{\bd{}also}~(4.01) & \tk{\bd{}offered}~(1.50) & \tk{\bd{}however}~(2.97) \\
16 & \tk{\bd{}make}~(5.64) & \tk{\bd{}supported}~(1.71) & \tk{\bd{}rather}~(4.24) & 41 & \tk{\bd{}that}~(4.00) & \tk{\bd{}whether}~(1.48) & \tk{\bd{}estimated}~(2.94) \\
17 & \tk{\bd{}find}~(5.58) & \tk{\bd{}raise}~(1.69) & \tk{\bd{}on}~(4.19) & 42 & \tk{\bd{}understand}~(3.98) & \tk{\bd{}setting}~(1.44) & \tk{\bd{}ensure}~(2.93) \\
18 & \tk{\bd{}said}~(5.57) & \tk{\bd{}affected}~(1.69) & \tk{\bd{}been}~(4.00) & 43 & \tk{\bd{}can}~(3.92) & \tk{\bd{}Although}~(1.44) & \tk{\bd{}face}~(2.84) \\
19 & \tk{\bd{}things}~(5.31) & \tk{\bd{}popularity}~(1.69) & \tk{\bd{}potential}~(3.84) & 44 & \tk{\bd{}problem}~(3.86) & \tk{\bd{}focus}~(1.42) & \tk{\bd{}its}~(2.84) \\
20 & \tk{\bd{}probably}~(5.26) & \tk{\bd{}achievement}~(1.67) & \tk{\bd{}over}~(3.68) & 45 & \tk{\bd{}wants}~(3.82) & \tk{\bd{}major}~(1.41) & \tk{\bd{}faced}~(2.84) \\
21 & \tk{\bd{}problems}~(5.24) & \tk{\bd{}persuade}~(1.67) & \tk{-}~(3.60) & 46 & \tk{\bd{}too}~(3.80) & \tk{However}~(1.40) & \tk{ening}~(2.84) \\
22 & \tk{\bd{}more}~(5.18) & \tk{\bd{}partly}~(1.65) & \tk{\bd{}launched}~(3.38) & 47 & \tk{\bd{}\bd{}}~(3.80) & \tk{\bd{}produced}~(1.38) & \tk{\bd{}tackle}~(2.84) \\
23 & \tk{\bd{}not}~(5.17) & \tk{\bd{}led}~(1.62) & \tk{\bd{}individuals}~(3.36) & 48 & \tk{\bd{}stop}~(3.80) & \tk{\bd{}incredibly}~(1.37) & \tk{\bd{}whose}~(2.83) \\
24 & \tk{\bd{}lots}~(4.99) & \tk{\bd{}relief}~(1.62) & \tk{\bd{}though}~(3.36) & 49 & \tk{\bd{}pay}~(3.78) & \tk{\bd{}suggest}~(1.37) & \tk{\bd{}the}~(2.82) \\
25 & \tk{.\textbackslash{}n}~(4.89) & \tk{\bd{}earning}~(1.62) & \tk{\bd{}along}~(3.34) & 50 & \tk{\bd{}She}~(3.76) & \tk{\bd{}extremely}~(1.37) & \tk{\bd{}case}~(2.82) \\
\bottomrule
\end{tabular}
\caption{Top-50 tokens by statistical score for \textsc{Ose}. \textbf{E}: elementary; \textbf{I}: intermediate; \textbf{A}: advanced. The leading dot (\bd{}) marks a token-internal word boundary.}
\label{tab:facevalid-ose}
\end{table*}

\clearpage
\onecolumn
\section{Prompts}
\label{app:prompt}
\newcommand{\promptnote}[1]{%
    {\color{gray}\texttt{#1}}\par
}
\subsection{Prompts Used for Prompt-only}
\begin{promptbox}[title=Prompt Templates]
You are a writing assistant that produces articles about a given subject. 
Hard constraints (must follow): \\
- Do not output any URLs, links, or domain names (examples: http, https, www, .com, .net, .org, .io, .ai). \\
- Output plain text only.\\

\promptnote{\# OSE dataset}
Paraphrase the following article. Write at an \{\texttt{elementary, intermediate, advanced}\} level. \\
Rewrite the following text in a different style while maintaining the same meaning: \{\texttt{corpus}\} \\
Rewritten version: \\

\promptnote{\# WikiPol dataset}
\textbf{Impolite:}\\
Rewrite the following text to be impolite (direct, curt, or somewhat rude).\\
\textbf{Polite:}\\
Rewrite the following text to be polite (respectful, considerate, with greetings or gratitude). \\
\textbf{Neutral:}\\
Rewrite the following text to be neutral (neither clearly polite nor impolite). \\ \\
Rewrite the following text while maintaining the same meaning: \{\texttt{text}\} \\
Rewritten version:

\end{promptbox}

\subsection{Prompts Used for Judge LLM}
\label{app:judge_prompt}


\begin{promptbox}[title=\textsc{WikiPol} Prompt]
You are annotating the PERCEIVED POLITENESS of a written request.\bigskip

Context:\\
- Judge how polite the request would SOUND to a typical recipient (perception), not the writer's intent.\\
- Do not infer sarcasm or hostility unless it is explicit in the wording.\bigskip

Step 1) Assign a continuous politeness score s in [-2, +2]\\
- +2 = very polite, +1 = somewhat polite, 0 = neither clearly polite nor clearly impolite,
  -1 = somewhat impolite, -2 = very impolite.\bigskip

Use these evidence cues (not strict rules; weigh them holistically):\bigskip

Politeness-increasing cues:\\
- Greetings / sign-offs\\
- Gratitude\\
- Apologies\\
- Deference / respect markers\\
- Indirectness or softening (hedges, could/would, conditional phrasing)\\
- ``please'' as a softener (especially sentence-medial)\bigskip

Politeness-decreasing cues:\\
- Direct imperatives / commands\\
- Obligation presuppositions (you need/must/should)\\
- Curt phrasing / lack of mitigation\\
- Negative/complaining tone or blaming language\\
- ``Please'' used as a command (sentence-initial)\bigskip

Step 2) Map the score to a 3-way label\\
- POLITE:   s $\ge$ +0.75\\
- IMPOLITE: s $\le$ -0.75\\
- NEUTRAL:  otherwise (mixed/weak cues)\bigskip

Output ONLY a JSON object:\\
\{``label'':``POLITE|IMPOLITE|NEUTRAL'',``confidence'':0-1,\\
 ``reasons'':[``$\ldots$'',``$\ldots$'',``$\ldots$''],\\
 ``quotes'':[``$\ldots$'',``$\ldots$'',``$\ldots$'']\}\bigskip

Rules for output:\\
- reasons: 1-3 items. Each reason MUST explicitly cite one cue name above\\
  (e.g., "Gratitude: expresses thanks", "Direct imperatives: command-like wording").\\
- quotes: 1-3 items, <= 5 words each, copied verbatim from the INPUT TEXT.\\
- Every reason must be supported by at least one quote; if not, omit that reason.\\
- confidence guide:\\
  0.85-1.00 = multiple strong cues align\\
  0.60-0.84 = some clear cues\\
  0.40-0.59 = weak/mixed cues (often NEUTRAL)
\end{promptbox}

\begin{promptbox}[title=\textsc{OSE} Prompt]
You are a strict evaluator for OneStopEnglish-style rewriting levels:\\
ELEMENTARY, INTERMEDIATE, ADVANCED.\bigskip

Core principle:\\
- Judge the WRITING STYLE (simplification vs journalistic compression), not the topic.\\
- Even ELEMENTARY may contain advanced topic words; do NOT up-level based on topic vocabulary.\bigskip

What to focus on:\\
A) Simplification signals (push toward ELEMENTARY)\\
- ``Spell-out'' paraphrases and definitions (e.g., X that does Y; ``called $\ldots$''; explaining terms)\\
- Sentence splitting: facts spread across many short/plain sentences\\
- Basic/local cohesion: heavy reliance on and/but/so/because; list-like sequencing\\
- Repetition / low variation: repeated frames, repeated key words\\
- More explicit moral/author commentary in simple wording\bigskip

B) Journalistic compression signals (push toward ADVANCED)\\
- Dense noun phrases and precise verbs (e.g., insists/denies/echoes/anticipates/deemed/bracing)\\
- Strong framing: setup $\to$ development $\to$ implications; effective transitions (however/nonetheless/whereas)\\
- Consistently natural collocations; little learner-like ``spell-out'' wording\\
- Information density: attribution, qualifiers, contrast, embedded clauses handled well across the text\bigskip

Text integrity rule (important):\\
- If the prose contains obvious corruption (truncated sentences, duplicated fragments inserted mid-sentence, scrambled ordering),\\
  treat this as NOISE. Do not automatically equate noise with low level.\\
  If corruption prevents reliable judging, choose the LOWER label and lower confidence.\bigskip

Definitions (use these exactly):\\
- ELEMENTARY: learner-simplified prose. Frequent spell-out paraphrases/definitions, short/plain sentences, basic connectors, repetition, and weaker global framing. Major grammar errors may occur but are NOT required.\\
- INTERMEDIATE: meaning is stable. Some complex sentences and connectors appear, but prose still feels simplified; fewer dense journalistic turns than ADVANCED; occasional awkwardness may remain.\\
- ADVANCED: consistently controlled, information-dense, journalistic-style prose. Compression/framing and precise word choice are sustained across the text. Minor typos are acceptable if they do not stand out.\bigskip

Decision procedure (must follow):\\
1) Ignore non-prose (code, templates, ``Continue writing'', markdown artifacts). Judge only meaningful prose.\\
2) Choose ELEMENTARY if at least TWO simplification signals (A) are strong and persistent across the text.\\
3) Otherwise choose ADVANCED if at least THREE compression signals (B) are strong and persistent across the text.\\
4) Otherwise choose INTERMEDIATE.\\
5) Tie-break: if unsure between two labels, choose the LOWER label.\bigskip

Output ONLY a JSON object:\\
\{``label'':``ELEMENTARY|INTERMEDIATE|ADVANCED'',``confidence'':0-1,\\
 ``reasons'':[``$\ldots$'',``$\ldots$'',``$\ldots$''],\\
 ``quotes'':[``$\ldots$'',``$\ldots$'',``$\ldots$'']\}\\
- Reasons must cite the signals above (simplification vs compression).\\
- Quotes: $\le$ 12 words each, copied verbatim from the text.
\end{promptbox}
\twocolumn

\end{document}